\documentclass[10pt,twocolumn,letterpaper]{article}

\usepackage{cvpr}              %

\usepackage{graphicx}
\usepackage{amsmath}
\usepackage{amssymb}
\usepackage{booktabs}
\usepackage{multirow}
\usepackage{pifont}
\usepackage{adjustbox}
\usepackage{enumitem}
\usepackage{array}
\usepackage{tabulary}
\usepackage{colortbl}
\usepackage[accsupp]{axessibility}
\usepackage{xcolor}

\usepackage[pagebackref,breaklinks,colorlinks]{hyperref}

\newcommand{\cmark}{\ding{51}}%
\newcommand{\xmark}{\ding{55}}%

\definecolor{darkgray}{RGB}{128,128,128}

\newcommand{\app}{\raise.17ex\hbox{$\scriptstyle\sim$}}

\newcolumntype{x}[1]{>{\centering\arraybackslash}p{#1pt}}

\newlength\savewidth
\newcommand{\tablestyle}[2]{\setlength{\tabcolsep}{#1}\renewcommand{\arraystretch}{#2}\centering\footnotesize}
\makeatletter\renewcommand\paragraph{\@startsection{paragraph}{4}{\z@}
  {.5em \@plus1ex \@minus.2ex}{-.5em}{\normalfont\normalsize\bfseries}}\makeatother

\usepackage[capitalize]{cleveref}
\crefname{section}{Sec.}{Secs.}
\Crefname{section}{Section}{Sections}
\Crefname{table}{Table}{Tables}
\crefname{table}{Tab.}{Tabs.}

\def\confName{CVPR}
\def\confYear{2023}

\begin{document}

\title{Content-aware Token Sharing for Efficient Semantic Segmentation \\ with Vision Transformers}

\author{Chenyang Lu$^{*}$, Daan de Geus$^{*}$, Gijs Dubbelman\\
Eindhoven University of Technology \\
{\tt\small \{c.lu.2, d.c.d.geus, g.dubbelman\}@tue.nl}
}

\maketitle

\def\thefootnote{*}\footnotetext{Both authors contributed equally.}\def\thefootnote{\arabic{footnote}}

\begin{abstract}

This paper introduces Content-aware Token Sharing (CTS), a token reduction approach that improves the computational efficiency of semantic segmentation networks that use Vision Transformers (ViTs). Existing works have proposed token reduction approaches to improve the efficiency of ViT-based image classification networks, but these methods are not directly applicable to semantic segmentation, which we address in this work. We observe that, for semantic segmentation, multiple image patches can share a token if they contain the same semantic class, as they contain redundant information. Our approach leverages this by employing an efficient, class-agnostic policy network that predicts if image patches contain the same semantic class, and lets them share a token if they do. With experiments, we explore the critical design choices of CTS and show its effectiveness on the ADE20K, Pascal Context and Cityscapes datasets, various ViT backbones, and different segmentation decoders. With Content-aware Token Sharing, we are able to reduce the number of processed tokens by up to 44\%, without diminishing the segmentation quality.

\end{abstract}

\section{Introduction}
\label{sec:intro}

In recent years, many works have proposed replacing Convolutional Neural Networks (CNNs) with Vision Transformers (ViTs)~\cite{dosovitskiy_image_2021} to solve various computer vision tasks, such as image classification~\cite{dosovitskiy_image_2021,liu_swin_2021,liu2022swinv2,bao2022beit}, object detection~\cite{liu_swin_2021,carion2020detr,zhu2021deformdetr} and semantic segmentation~\cite{strudel_segmenter_2021,xie2021segformer,cheng2022mask2former,liu_swin_2021,zheng2021setr,ranftl2021dpt}. ViTs\footnote{In this work we use the term \textit{ViTs} for the complete family of vision transformers that purely apply global self-attention.} apply multiple layers of multi-head self-attention~\cite{vaswani_attention_2017} to a set of tokens generated from fixed-size image patches. ViT-based models now achieve state-of-the-art results for various tasks, and it is found that ViTs are especially well-suited for pre-training on large datasets~\cite{bao2022beit,he2022mae}, which in turn yields significant improvements on downstream tasks. However, the use of global self-attention, which is key in achieving good results, means that the computational complexity of the model is quadratic with respect to the input tokens. As a result, these models become particularly inefficient for dense tasks like semantic segmentation, which are typically applied to larger images than for image classification.

\begin{figure}[!t]
    \centering
    \includegraphics[width=\linewidth]{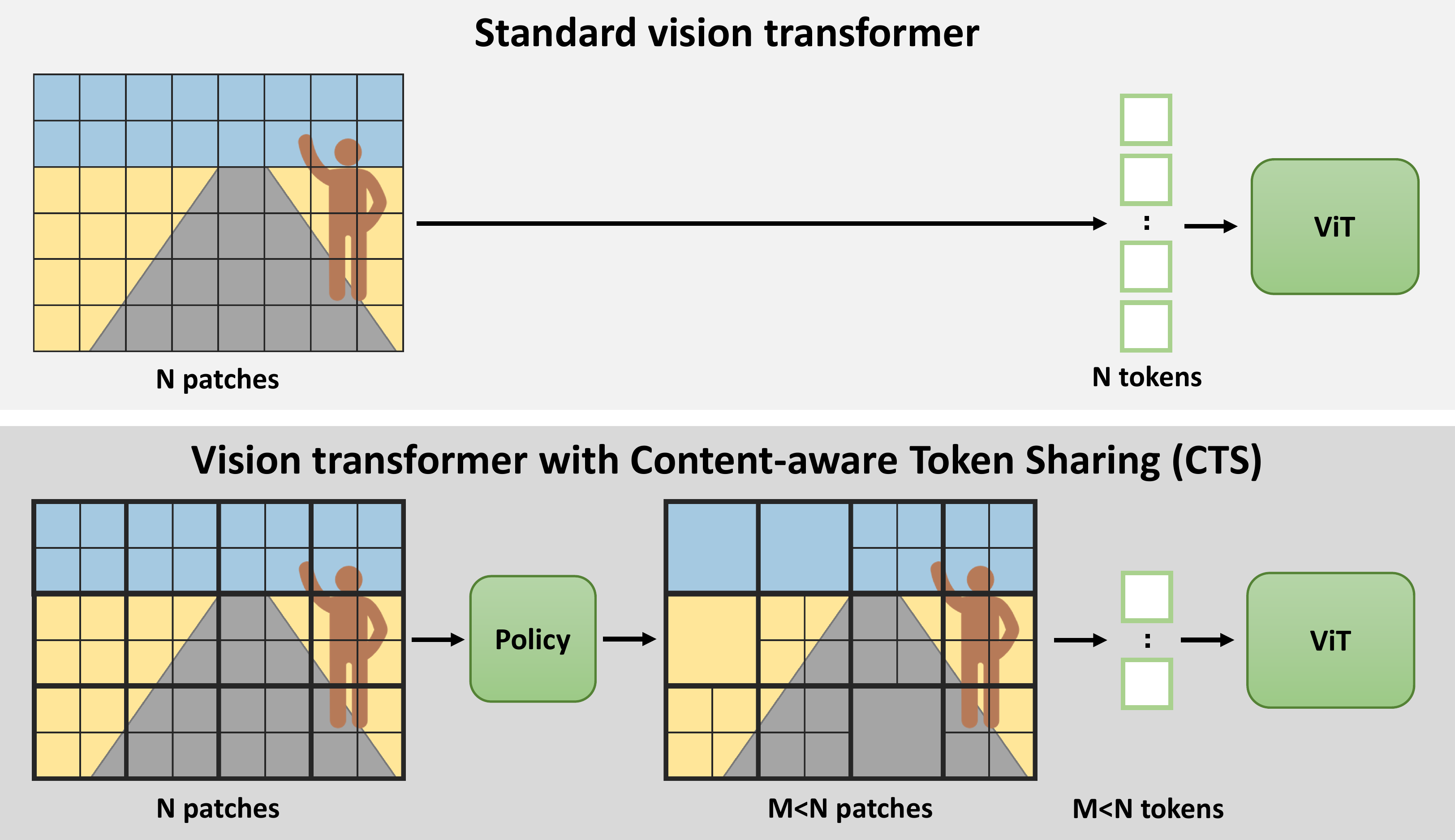}
     \caption{\textbf{Content-aware token sharing (CTS).} Standard ViT-based segmentation networks turn fixed-size patches into tokens, and process all of them. To improve efficiency, we propose to let semantically similar patches share a token, and achieve considerable efficiency boosts without decreasing the segmentation quality.}
    \label{fig:eye_catcher}
    \vspace{-10pt}
\end{figure}

Recent works address these efficiency concerns in two different ways. Some works propose new ViT-based architectures to improve efficiency, which either make a combination of global and local attention~\cite{liu_swin_2021,focalvit2021} or introduce a pyramid-like structure inspired by CNNs~\cite{pyramidvit2021,liu_swin_2021,graham2021levit}. Alternatively, to reduce the burden of the quadratic complexity, some works aim to reduce number of tokens that are processed by the network, by either discarding~\cite{rao2021dynamicvit,tang2022patchslimming,yin_avit_2022} or merging~\cite{renggli2022tokenmerger,liang2022evit,bolya2022tome} the least relevant tokens. These works, which all address the image classification task, find that a similar accuracy can be achieved when taking into account only a subset of all tokens, thereby improving efficiency.
However, these methods, which are discussed further in \Cref{sec:related_work}, are not directly applicable to semantic segmentation. First, we cannot simply discard certain tokens, as each token represents an image region for which the semantic segmentation task requires predictions. Second, existing token merging approaches allow any combination of tokens to be merged, through multiple stages of the network. As a result, `unmerging' and spatial reorganization of tokens, which is necessary because semantic segmentation requires a prediction for each original token, is non-trivial.

In this work, we present a much simpler, yet highly effective and generally applicable token reduction approach for ViT-based semantic segmentation networks. The goal of this approach is to improve the efficiency without decreasing the segmentation quality. Token reduction methods for image classification already found that merging redundant tokens only results in a limited accuracy drop~\cite{renggli2022tokenmerger,liang2022evit,bolya2022tome}. For semantic segmentation, we observe that there are many neighboring patches that contain the same semantic class, and that the information contained by these patches is likely redundant. Therefore, we hypothesize that neighboring patches containing the same class can share a token without negatively impacting the segmentation quality of a semantic segmentation network. To leverage this, we propose an approach that reduces tokens by (1) using a policy that identifies which patches can share a token before any tokens enter the ViT, and (2) grouping only rectangular neighboring regions, which allows for straightforward reassembling of tokens at the output of the ViT backbone. The main challenge that we tackle is to find this policy without introducing a heavy computational burden.

We propose to tackle this policy challenge by turning the problem into a simple binary classification task that can be solved by a highly efficient CNN model, our \textit{policy network}. Concretely, it predicts whether 2$\times$2 neighboring patches contain the same class, and lets these patches share a token if they do. Our complete approach, which we call \textit{Content-aware Token Sharing} (CTS), first applies this policy network, then lets patches share tokens according to the predicted policy, feeds the reduced set of tokens through the ViT, and finally makes a semantic segmentation prediction using these tokens (see \Cref{sec:method}). CTS has several advantages by design. First, it does not require any modifications to the ViT architecture or its training strategy for pre-training or fine-tuning. As a result, it is compatible with all backbones that purely use global self-attention, as well as any semantic segmentation decoder or advanced pre-training strategy, \eg, BEiT~\cite{bao2022beit}. Second, by reducing the number of tokens before inputting them to the ViT, the efficiency improvement is larger than when token reduction is applied gradually in successive stages of the ViT, as done in existing methods for image classification~\cite{bolya2022tome,yin_avit_2022}. Both advantages are facilitated by our policy network, which is trained separately from the ViT, and is so efficient that it only introduces a marginal computational overhead for large ViTs applied to high-resolution images.

With experiments detailed in \Cref{sec:exp}, we show that our CTS approach can reduce the total number of processed tokens by at least 30\% without decreasing the segmentation quality, for multiple datasets. We also show that this holds for transformer backbones of many different sizes, initialized with different pre-trained weights, and using various segmentation decoders. For more detailed results, we refer to \Cref{sec:results}. With this work, we aim to provide a foundation for future research on efficient transformer architectures for per-pixel prediction tasks. The code for this work is available through \url{https://tue-mps.github.io/CTS}.

To summarize, the contributions of this work are:
\begin{itemize}[noitemsep,nolistsep]
    \item A generally applicable token sharing framework that lets semantically similar neighboring image patches share a token, improving the efficiency of ViT-based semantic segmentation networks without reducing the segmentation quality.
    \item A content-aware token sharing policy network that efficiently classifies whether a set of neighboring patches should share a token or not.
    \item A comprehensive set of experiments with which we show the effectiveness of the proposed CTS approach on the ADE20K~\cite{zhou_scene_2017}, Pascal Context~\cite{mottaghi_role_2014}, and Cityscapes~\cite{cordts_cityscapes_2016} datasets, for a wide range of different ViTs and decoders.  
\end{itemize}

\section{Related work}
\label{sec:related_work}

\textbf{Efficient vision transformers.} Several works aim to mitigate the effect of the quadratic complexity of vision transformers, and make them more efficient. 
As mentioned in the introduction, different strategies exist to improve the efficiency of ViTs \cite{liu_swin_2021,focalvit2021,pyramidvit2021,graham2021levit}, and our CTS falls in the category of approaches that aim to reduce tokens. This can be achieved by either discarding~\cite{yin_avit_2022,rao2021dynamicvit,tang2022patchslimming} or merging tokens~\cite{renggli2022tokenmerger,bolya2022tome,liang2022evit,ryoo_tokenlearner_2022} at different stages of the network, and the number of processed tokens per image frequently depends on the complexity of the image. Each of these methods, which are strictly proposed for image classification, makes the decision to discard or merge tokens in a different way, \eg, based on the similarity of tokens in embedding space~\cite{bolya2022tome}, the attentiveness of tokens~\cite{liang2022evit} or with learned parameters~\cite{renggli2022tokenmerger}. To be even more adaptive, AdaViT~\cite{meng2022adavit} and V-MoE~\cite{riquelme2021vmoe} also introduce sparsity in the network architecture, by allowing tokens or entire images to be processed by a subset of the parameters, and DVT~\cite{wang_not_2021} varies the processed patch size based on the confidence of the network's predictions. The findings of these works are similar: it is possible to reduce a significant number of tokens with little impact on the image classification accuracy, implying that many tokens are redundant.

\textbf{Semantic segmentation with vision transformers.} 
Vision transformers are also commonly applied to semantic segmentation, usually in one of two different ways. First, there are works that propose custom transformer architectures that solve the entire task, for semantic segmentation~\cite{xie2021segformer,zheng2021setr,wang2022rtformer} but also for more complex tasks like panoptic segmentation~\cite{kirillov_panoptic_2019,li2022pansegformer,wang2021maxdeeplab} and part-aware panoptic segmentation~\cite{de_geus_part-aware_2021,li2022panpartformer}. Second, more commonly, works propose improvements to either a transformer-based backbone~\cite{liu_swin_2021,pyramidvit2021,gu2022hrvit,dong2022cswin,chu2021twins} or task-specific decoder~\cite{strudel_segmenter_2021,cheng2022maskformer,cheng2022mask2former,zhang2021knet,ranftl2021dpt}. An advantage of a modular backbone--decoder architecture is that backbones can simply be replaced by ones that are pre-trained on more data. This is important because it has been shown that pre-training transformers on large -- sometimes even multimodal -- datasets~\cite{he2022mae,liu_swin_2021,steiner2022trainvit,bao2022beit,chen2022vitadapter} leads to significant improvements on downstream tasks such as semantic segmentation. Being compatible with such advanced pre-training methods is therefore an important property of our CTS.

Several of the proposed works also focus on improving efficiency, usually by proposing architecture changes~\cite{liu_swin_2021,wang2022rtformer,pyramidvit2021,zhang2022topformer,mehta2022mobilevit}. However, to the best of our knowledge, the findings from the field of image classification that certain tokens are redundant, have not been applied to semantic segmentation. In this work, inspired by these findings, we propose an approach that allows neighboring image patches to \textit{share} a token when they have similar content, to improve the efficiency while maintaining the segmentation quality.

\textbf{Efficient semantic segmentation.} Many works focus purely on  efficient semantic segmentation~\cite{zhao2018icnet,yo2018bisenet,zhang2022topformer,mehta2022mobilevit}. However, these works generally propose architectures that are purely optimized for efficiency, and achieve a segmentation quality that is below the state of the art. In contrast, our approach is designed to improve the efficiency of any existing or future state-of-the-art ViT-based segmentation network, without compromising the segmentation quality.

\section{Content-aware token sharing}
\label{sec:method}

\begin{figure}[!t]
    \centering
    \includegraphics[width=\linewidth]{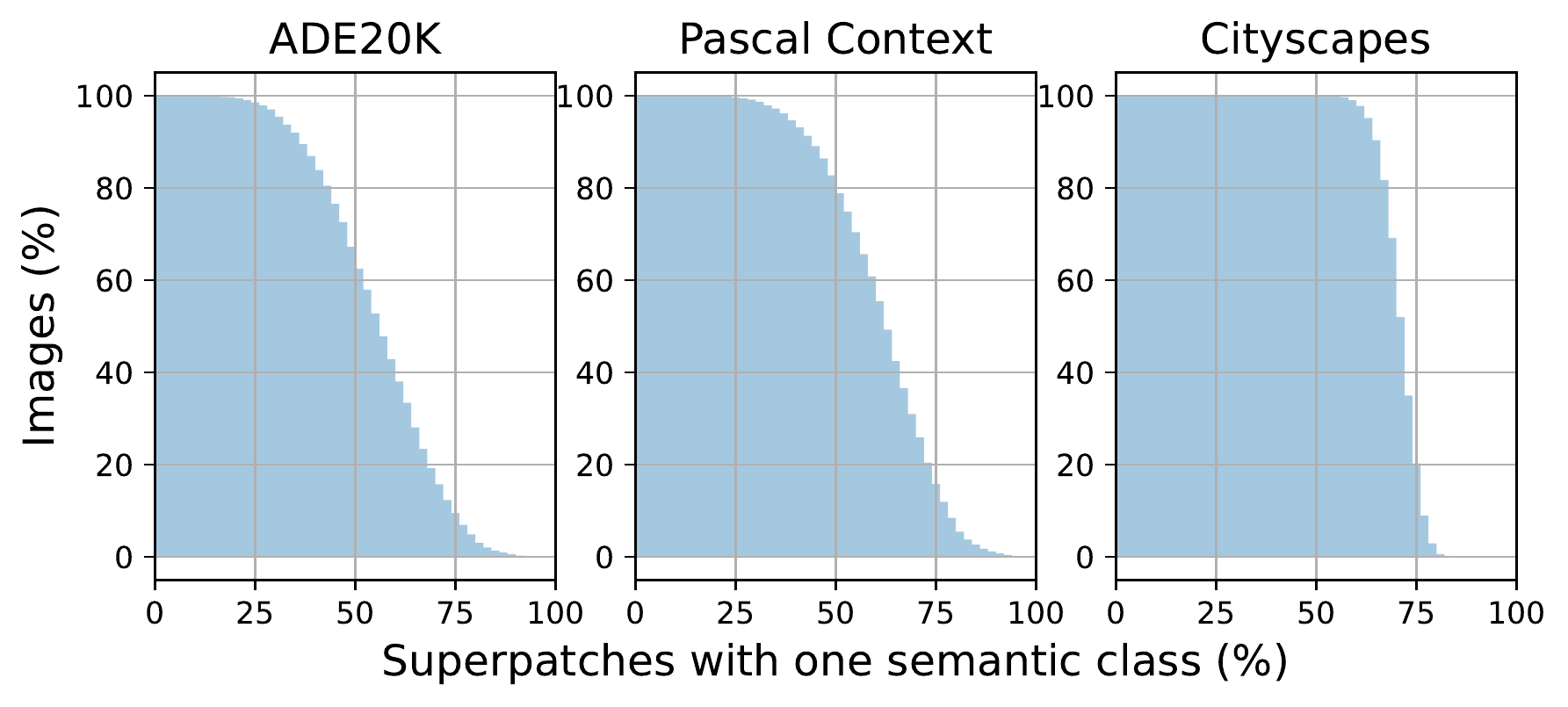}
    \vspace{-20pt}
    \caption{\textbf{Dataset statistics.} We show how many images have a certain percentage of superpatches that contain a single semantic class. We hypothesize that these superpatches can share a token.}
    \label{fig:patch_histograms}
    \vspace{-10pt}
\end{figure}

\subsection{Problem definition}
\label{sec:method_problem}
In ViT-based segmentation architectures, an input RGB image $I \in \mathbb{R}^{3\times{\text{H}}\times{\text{W}}}$ is partitioned into a set of $\text{N}$ image patches $\mathcal{I} = \{I_1, ...,I_\text{N}\}$. These image patches are linearly projected into a set of tokens $\mathcal{T} = \{T_1,...,T_\text{N}\}$, added to their corresponding learnable positional embeddings, and are subsequently fed into transformer blocks~\cite{dosovitskiy_image_2021}. The total number of tokens $\text{N}$ depends on the size of the input image and the pre-defined patch size. A transformer-based backbone $f: \mathcal{T} \mapsto \mathcal{L}$ maps input tokens $\mathcal{T}$ into a set of predicted tokens $\mathcal{L} = \{L_1,...,L_\text{N}\}$ containing semantic information. The transformer model can have any architecture applying purely global self-attention, \eg, ~\cite{dosovitskiy_image_2021,bao2022beit}. The next step is to feed the output tokens $\mathcal{L}$ into a decoder $g$. There are two options for this decoder. (1) If the decoder requires tokens, \eg, a transformer decoder~\cite{strudel_segmenter_2021}, the tokens are immediately fed through the decoder $g: \mathcal{L} \mapsto \mathcal{O}_{tkn}$, where $\mathcal{O}_{tkn} = \{O_1,...,O_\text{N}\}$ is a set of per-token semantic predictions. These predictions are then rearranged and upsampled to per-pixel segmentation predictions $O \in \mathbb{R}^{\text{C}\times{\text{H}}\times{\text{W}}}$ for $\text{C}$ classes. (2) If the decoder requires spatially organized features, \eg, a CNN-based segmentation head~\cite{xiao2018unified}, the tokens are first rearranged based on their original position to get features $L_{spat} \in \mathbb{R}^{\text{E}\times{\text{H}_{\text{L}}}\times{\text{W}_{\text{L}}}}$, where $\text{E}$ is the embedding dimension and $\text{H}_{\text{L}},\text{W}_{\text{L}}$ are fractions of the original $\text{H,W}$. These are then fed through the segmentation decoder $g: {L}_{spat} \mapsto O$ to obtain the segmentation predictions.

\textbf{The problem} that is tackled by our approach is decreasing the size of $\mathcal{T}$ to reduce computation, by identifying patches in $\mathcal{I}$ that can share a token, without diminishing the segmentation quality of $O = g(f(\mathcal{T}))$.

\subsection{Content-aware token sharing framework} 
\label{sec:method_framwork}
Our solution is based on the observation that some patches contain redundant information. In Figure \ref{fig:patch_histograms}, we show for several datasets the number of images that have a specific percentage of 2$\times$2 neighboring image patches, \ie, \textit{superpatches}, in $\mathcal{I}$ that contain only one semantic class. We observe that in all images from all these datasets, roughly 25\% of superpatches contain only one semantic class, and for many images this percentage is even higher. From these statistics, it is clear that there are many semantically uniform regions, and we hypothesize that efficiency can be improved without diminishing segmentation quality by processing these patches jointly. To achieve this, we propose \textit{Content-aware Token Sharing} (CTS), which first predicts what superpatches in $\mathcal{I}$ contain only one semantic class, and then lets each of these sets of 2$\times$2 neighboring patches share the same, single token.

\begin{figure}[!t]
    \centering
    \includegraphics[width=\linewidth]{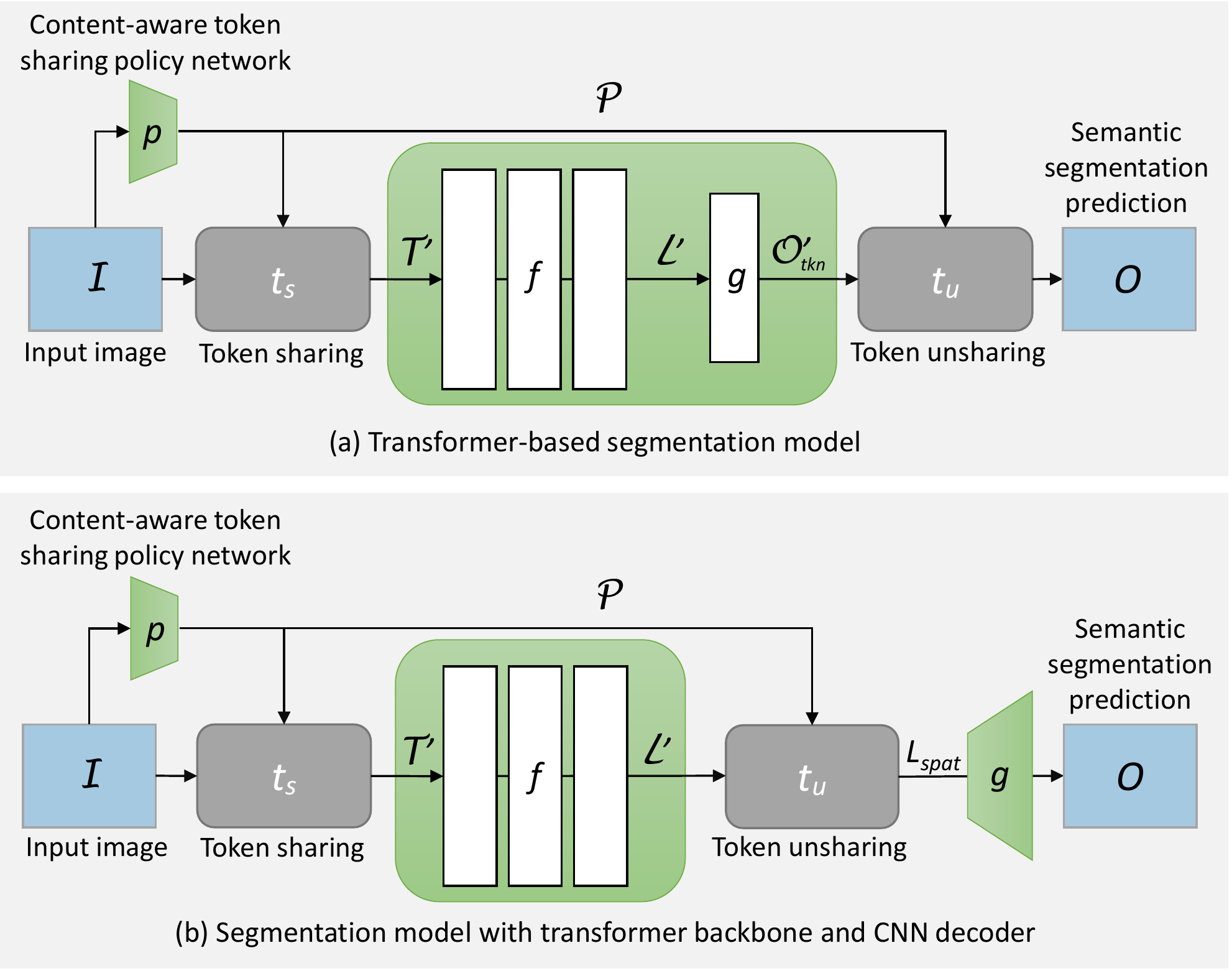}
    \vspace{-15pt}
    \caption{\textbf{Method overview.} We introduce a policy network $p$ that predicts what image patches can share a token without decreasing the performance. These patches are then combined into a single token using token sharing module $t_s$. Subsequently, remaining tokens are fed through the transformer model, and the output tokens are `unshared' using module $t_u$, either (a) after the per-token predictions are made or (b) before the per-pixel predictions are made.}

    \label{fig:main_diagram}
    \vspace{-15pt}
\end{figure}

To enable CTS for any conventional transformer-based model, we introduce a \textit{token sharing} function $t_{s}$, a \textit{token unsharing} function $t_{u}$, and a \textit{policy} model $p$. All these are implemented in a differentiable manner, such that end-to-end training can be performed as usual. 

The token sharing function $t_{s}$ maps the original image patches $\mathcal{I}$
and their positional embeddings 
into a new set of patches $\mathcal{I}^\prime = \{I^\prime_1,...,I^\prime_\text{M}\}$ with $\text{M}<\text{N}$ samples, with corresponding positional embeddings. This mapping is performed according to a predicted policy $\mathcal{P}$ that states which superpatches should share a single token, and which should have a unique token for each of their 2$\times$2 patches, as conventional.
This policy is predicted by the policy model $p$, which is explained in detail in \Cref{sec:method_policynet}. 
The reduced set of image patches $\mathcal{I}^\prime$ is then projected to a reduced set of tokens $\mathcal{T}^\prime$ to which conventional transformer-based backbones can be deployed without any modifications. This token sharing function can be formalized as 
$\mathcal{T}^\prime = {t_{s}}(\mathcal{I}, \mathcal{P}).$ In this function, each superpatch that can share a token according to policy $\mathcal{P}$ is bilinearly downsampled to the dimensions of a single patch, and linearly projected into a single token. As a result, the 2$\times$2 patches in this superpatch share the same token.

Given $\mathcal{T}^\prime$, the transformer-based backbone $f$ produces a set $\mathcal{L}^\prime = \{L^\prime_1,...,L^\prime_\text{M}\}$ of $\text{M}$ predicted tokens with semantic information. 
Then, again, there are two options for the decoder.
(1) When the set $\mathcal{L}^\prime$ including shared tokens is directly fed to the decoder, the resulting per-token predictions $\mathcal{O}^\prime_{tkn}$ need to be `unshared'. We use the policy $\mathcal{P}$ to identify the predictions that are made for shared tokens, simply bilinearly upsample them and divide them back into individual predictions, before reorganizing all tokens spatially and upsampling them to yield the complete semantic segmentation predictions $O$. This is formalized in Eq.~\textcolor{red}{1} and visualized in \Cref{fig:main_diagram}\textcolor{red}{a}.
(2) When the tokens in $\mathcal{L}^\prime$ should first be rearranged to features $L_{spat}$ before entering the decoder, $t_{u}$ first unshares the shared tokens by bilinearly upsampling and rearranging them to produce $\mathcal{L}$, before organizing $\mathcal{L}$ spatially as features $L_{spat}$ and feeding this through the decoder $g$, which outputs segmentation predictions $O$. This is formalized in Eq.~\textcolor{red}{2} and visualized in \Cref{fig:main_diagram}\textcolor{red}{b}.

\begin{equation*}
\begin{aligned}
&  \mathcal{P} = p(I);                                    &&  &&  &&  &&  \mathcal{P} = p(I); &&  \\
&  \mathcal{T}^\prime = t_{s}(\mathcal{I}, \mathcal{P});  && (1) &&  &&  &&  \mathcal{T}^\prime = t_{s}(\mathcal{I}, \mathcal{P}); && (2) \\
&  \mathcal{O}_{tkn}^\prime = g(f(\mathcal{T}^\prime)); && &&  && &&  \mathcal{L}^\prime =  f(\mathcal{T}^\prime); && \\
&  O = t_{u}(\mathcal{O}_{tkn}^\prime, \mathcal{P}).    &&  &&  && && O = g(t_{u}(\mathcal{L}^\prime, \mathcal{P})). &&  \\
\end{aligned}
\label{eq:main}
\end{equation*}

Our framework is compatible with many different backbones, decoders, and policy models. This flexibility, as well as the consistent efficiency gains and quality preservation, is demonstrated extensively in Section~\ref{sec:results}.

\subsection{Content-aware token sharing policy}
\label{sec:method_policynet}

The token sharing function $t_{s}$ that lets patches share tokens relies on the output $\mathcal{P}$ of a content-aware token sharing policy model $p$ that predicts whether a superpatch (containing 2$\times$2 neighboring patches) can share the same token without diminishing segmentation quality. Realizing such a CTS policy is crucial for our framework, but non-trivial. Given our hypothesis that patches containing the same semantic class do not require separate, unique tokens, we propose to solve this problem by training and using a highly efficient CNN model to predict if a superpatch contains a single semantic class, without explicitly predicting the classes. If a superpatch indeed contains a single class, its patches can share a single token. This class-agnostic model, defined as $\mathcal{P} = p(I)$, can be implemented with any lightweight network, and simply performs binary classification on each superpatch region in an input image. From the output of the network, the S highest-scoring superpatches are selected to share a token, to end up with $\text{M} = \text{N} - 3\text{S}$ tokens that are processed by the transformer-based network.

\begin{figure}[!t]
    \centering
    \includegraphics[width=0.24\linewidth]{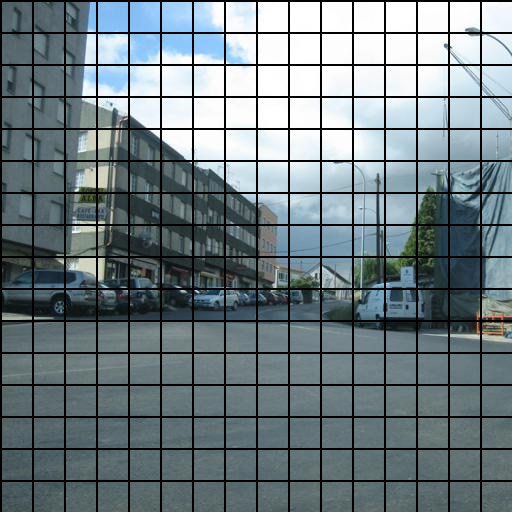}
    \includegraphics[width=0.24\linewidth]{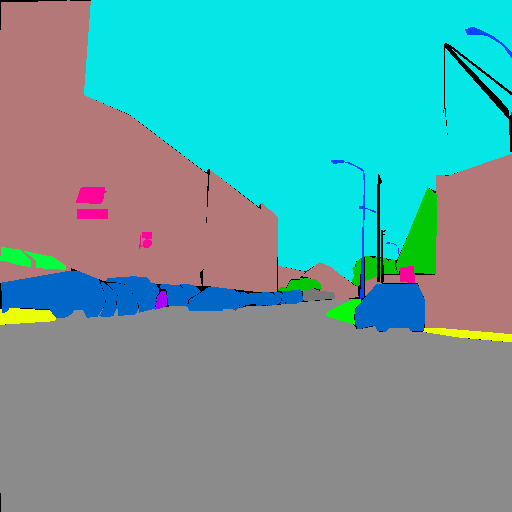}
    \includegraphics[width=0.24\linewidth]{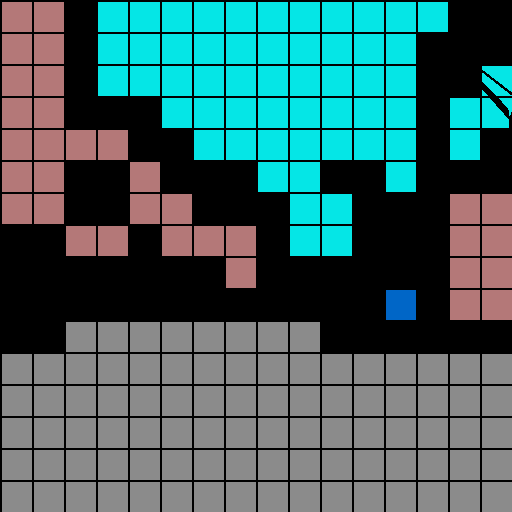} 
    \includegraphics[width=0.24\linewidth]{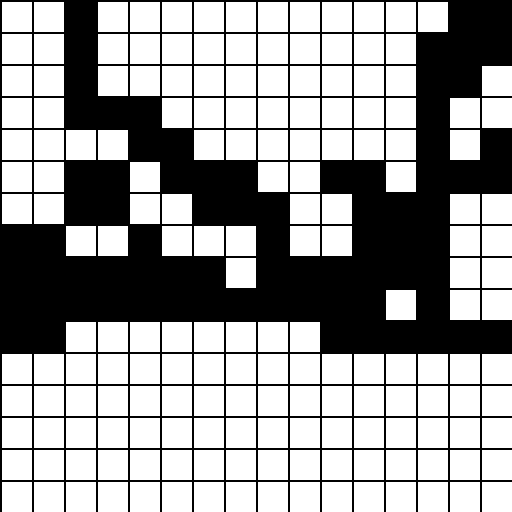} 
    \\
    \includegraphics[width=0.24\linewidth]{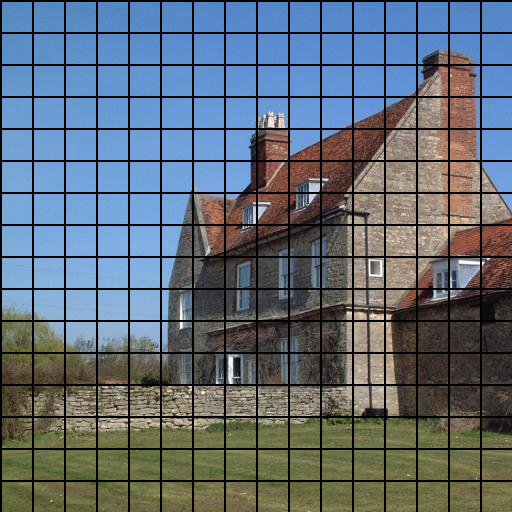}
    \includegraphics[width=0.24\linewidth]{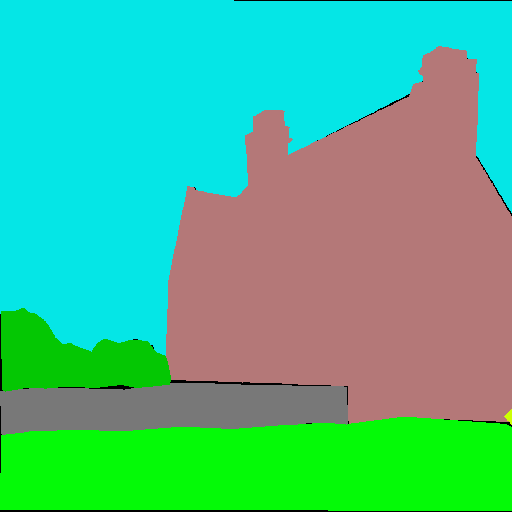}
    \includegraphics[width=0.24\linewidth]{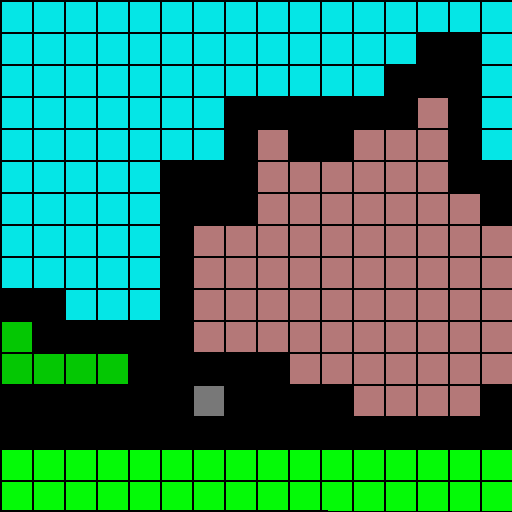} 
    \includegraphics[width=0.24\linewidth]{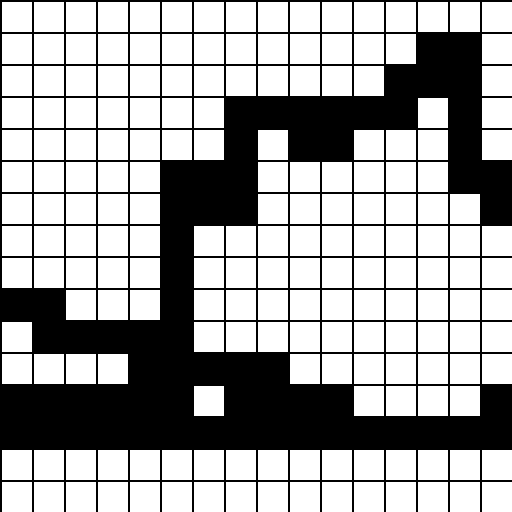}\\
    \includegraphics[width=0.24\linewidth]{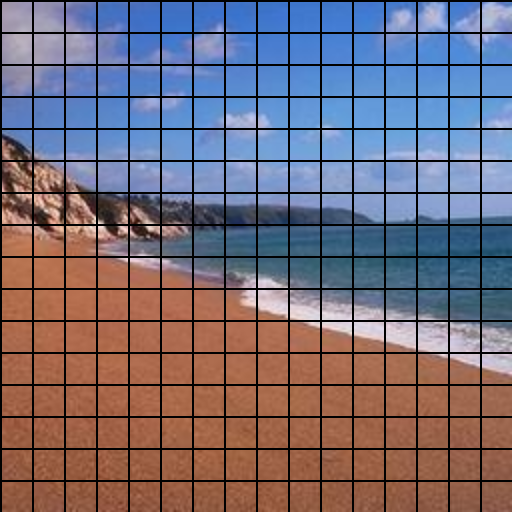}
    \includegraphics[width=0.24\linewidth]{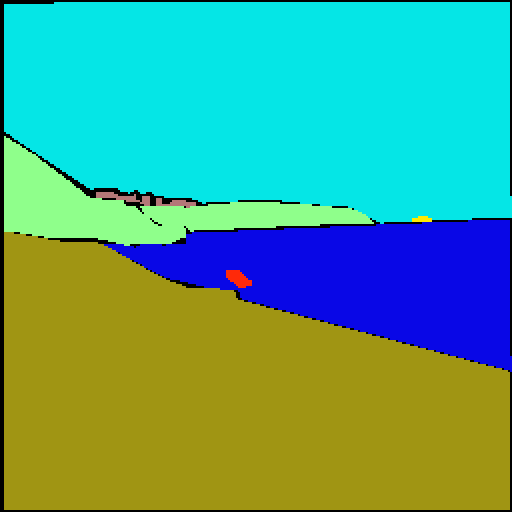}
    \includegraphics[width=0.24\linewidth]{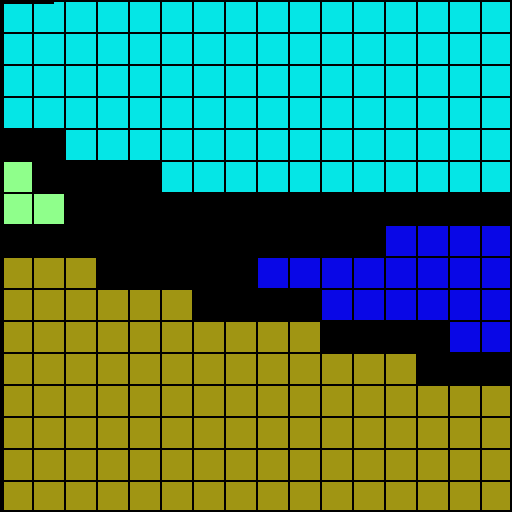} 
    \includegraphics[width=0.24\linewidth]{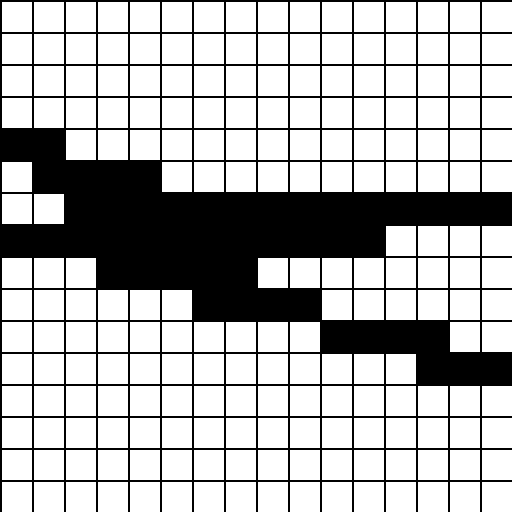}\\
    \caption{\textbf{Token sharing policy.} We teach our token sharing policy network that a superpatch should share a token if it contains a single semantic class. From left to right: (a) input image with grid of superpatches; (b) segmentation ground truth; (c) superpatches containing a single class; (d) class-agnostic policy ground truth. }
    \label{fig:grouping_examples}
    \vspace{-10pt}
\end{figure}

Our class-agnostic policy model is trained using a ground truth that is conveniently generated from the available semantic segmentation annotations.
Concretely, for each superpatch in an image, we analyze how many classes it contains in the semantic segmentation ground truth. If there is just a single semantic class, the ground truth for the policy model $p$ is set to true, otherwise it is false. This is visualized in \Cref{fig:grouping_examples}. Using this ground truth, the policy model is trained with a standard cross-entropy loss. 

It is possible to train our policy network in such a simple and efficient way because of the available annotations for this task. Unlike image classification and object detection annotations, which do not have sufficient granularity to directly determine what image regions are relevant for processing, semantic segmentation annotations provide labels for each pixel. This allows us to directly translate our hypothesis about redundancy of semantically similar patches into a binary ground truth per superpatch. Moreover, the fact that our policy model is class-agnostic is key to its efficiency. Because the binary task is significantly less complex than the main multi-class segmentation task, it can be performed by a highly efficient model, which is necessary to preserve the efficiency that is gained by reducing the number of tokens. In \Cref{sec:results}, we show that our class-agnostic approach outperforms a trivial solution that uses an auxiliary semantic segmentation prediction to obtain the semantically uniform superpatches.

\section{Experimental setup}
\label{sec:exp}

\subsection{Datasets}
To show the wide applicability of CTS for semantic segmentation, we conduct experiments on three datasets:  ADE20K \cite{zhou_scene_2017}, Pascal Context \cite{mottaghi_role_2014}, and Cityscapes \cite{cordts_cityscapes_2016}. Following most recent works, we use ADE20K for our main experiments and show the performance on the other two datasets to verify the general applicability of CTS.

\subsection{Metrics} We use the conventional \textit{mean intersection-over-union} (mIoU) to evaluate the segmentation performance. Since our work aims to improve the efficiency of segmentation networks, we also evaluate the throughput of each network, in terms of \textit{images/second} (Im/sec). To measure this, we test each model on an Nvidia A100 GPU and report the average throughput over 100 iterations after 50 warm-up iterations, using batches of 32 image crops of the same resolutions used for training. For the proposed policy network, we additionally evaluate the \textit{precision} of the predictions which superpatches should share a token. We choose precision because the ground truth can have a different number of shared superpatches for each image, while CTS always picks a fixed number of S superpatches to be shared.

\subsection{Implementation details} 

\textbf{Policy network.} The content-aware token sharing policy network is trained separately from the semantic segmentation network. Our default policy network is an EfficientNet-Lite0 model~\cite{tan_efficientnet_2019}, pre-trained on ImageNet-1K~\cite{russakovsky_imagenet_2015}. For all three datasets, we train the policy network for 50k iterations with batches of 8 images, using the AdamW \cite{loshchilov2018decoupled} optimizer. The learning rate is $5\times10^{-5}$ and weight decay is $10^{-3}$. During training, we apply the same resizing and cropping strategies used for the corresponding segmentation models.

\textbf{Segmentation networks.} Once the policy network is trained, we integrate it in the segmentation model and run the whole model end-to-end (see \Cref{fig:main_diagram}). Specifically, we feed the training images through the policy network, apply the resulting policy to share tokens, and train the segmentation network on the resulting tokens. To train the segmentation networks, we follow the original training strategies, to allow for fair comparisons. For Segmenter~\cite{strudel_segmenter_2021}, we optimize the networks with Stochastic Gradient Descent. For the ADE20K and Pascal Context datasets, the initial learning rate is set to 0.001, while for Cityscapes the initial learning rate is 0.01. All Segmenter experiments use polynomial learning rate decay, following the original settings~\cite{strudel_segmenter_2021}. The batch size is 16 for Pascal Context and 8 for the other datasets. For experiments with the UPerNet decoder~\cite{xiao2018unified}, we add our CTS to publicly available BEiT+UPerNet and ViT+UPerNet implementations~\cite{mmseg2020} and use the original training and evaluation settings. All backbones except for BEiT are pre-trained on ImageNet-1K~\cite{russakovsky_imagenet_2015,steiner2022trainvit}. For more details, see \Cref{sec:app:impl_details}.

\section{Experimental results}
\label{sec:results}

\subsection{Content-aware token sharing}
To show the effectiveness of the proposed content-aware token sharing (CTS) approach, we apply it to ViT-based segmentation network Segmenter~\cite{strudel_segmenter_2021}. In \Cref{tab:ade20k_grouping_schedules}, we show the results for different values of S, \ie, different token sharing settings, leading to different token reductions and efficiency boosts. When $\text{S}=0$, and no superpatches share a token, this is equal to the standard Segmenter~\cite{strudel_segmenter_2021}. When $\text{S}=\text{M}=256$, all superpatches share a token, leading to a token reduction of 75\%. 
From the results, we can see that reducing the number of tokens leads to an increase in throughput, as expected. Moreover, the results show that with CTS, up to 30\% of tokens can be reduced without compromising the segmentation quality, with a mIoU of 45.1 compared to a baseline of 45.0. In a naive token reduction setting, however, when random superpatches are selected to be shared, the mIoU is significantly worse, \ie, 42.9. This confirms our motivation that specific superpatches can share a token without diminishing the segmentation quality, if they are selected carefully, as with our CTS. And although random token sharing is faster than our CTS, due to the overhead introduced by the policy network, CTS still achieves a significant throughput increase of 33\% while keeping the segmentation quality the same.

\begin{figure}[t]
    \centering
    \includegraphics[width=\linewidth]{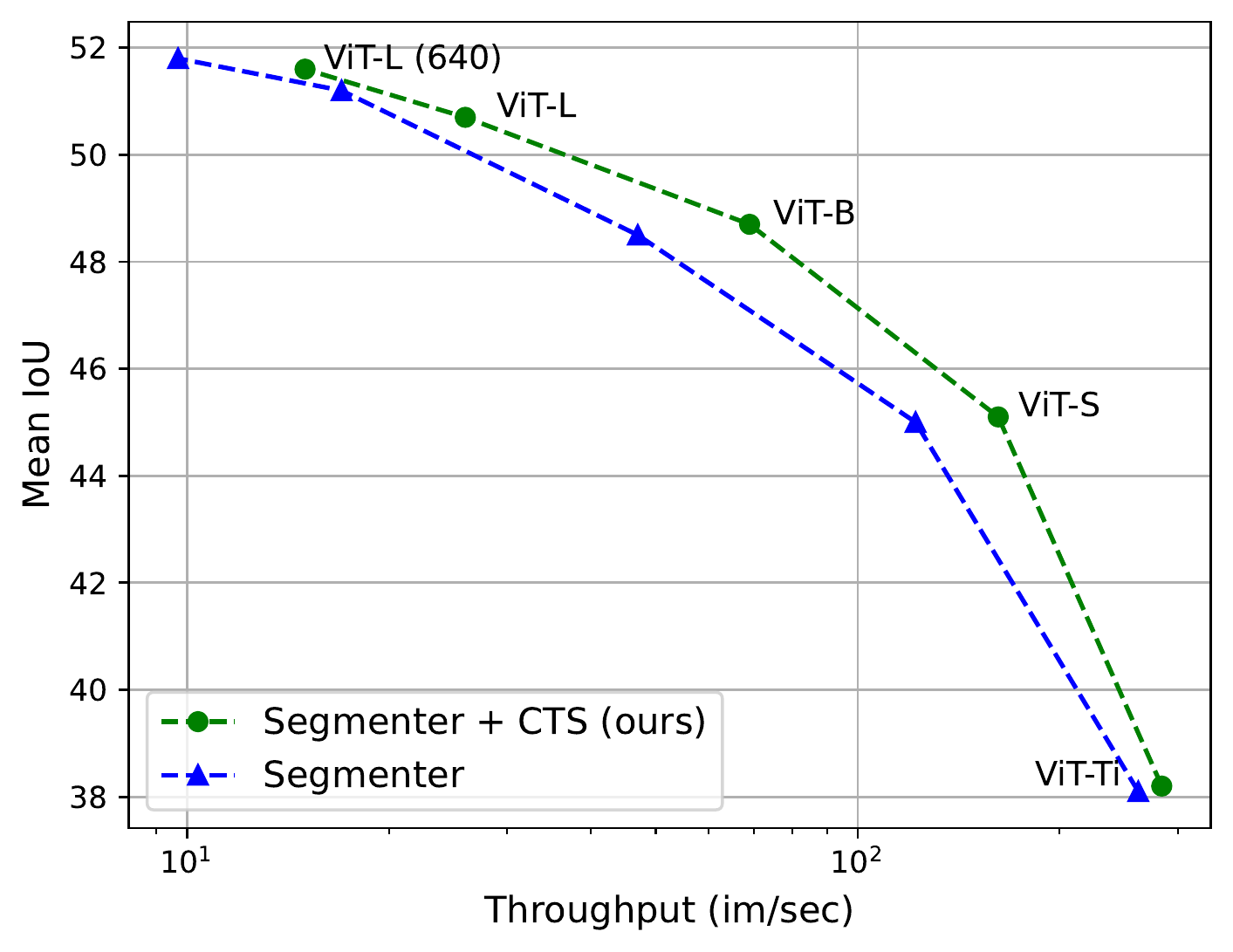}
    \vspace{-20pt}
     \caption{\textbf{Throughput vs.~mean IoU.} The efficiency and accuracy of Segmenter~\cite{strudel_segmenter_2021} with and without our content-aware token sharing approach. Results on the ADE20K val set~\cite{zhou_scene_2017}.}
    \label{fig:ade20k_backbones}
\end{figure}

With the identified optimal token reduction setting of 30\% for ADE20K, we also apply our CTS approach to Segmenter with larger and smaller backbones. From the results, depicted in \Cref{fig:ade20k_backbones}, it is immediately clear that CTS effectively improves the throughput for all backbone sizes by significantly reducing the number of tokens, while delivering a similar semantic segmentation performance. This further validates our approach.

\begin{table}[t]
  \centering
  \adjustbox{width=\linewidth}{
  \begin{tabular}{@{}ccccccc@{}}
    \toprule
    \multicolumn{3}{c}{\textbf{Tokens}} & \multicolumn{2}{c}{\textbf{CTS}} & \multicolumn{2}{c}{\textbf{Random sharing}} \\
    \cmidrule(l){1-3} \cmidrule(l){4-5} \cmidrule(l){6-7}
    Shared (S) & Total (M) & Reduction & mIoU & Im/sec & mIoU & Im/sec \\ 
    \midrule
    0 & 1024 & 0\% & 45.0 & 122 & - & - \\
    31 & 931 & 9\% & 45.3 & 123 & 44.7 & 139 \\
    41 & 901 & 12\% & 44.9 & 126 & 44.5 & 143 \\
    79 & 787  & 23\% & 45.0 & 146 & 43.6 & 168 \\
    \rowcolor{lightgray}
    103 & 715 & 30\% & 45.1 & 162 & 42.9 & 191 \\
    131 & 631  & 38\% & 44.7 & 180 & 42.4 & 216 \\
    156 & 556 & 46\% & 44.4 & 198 & 42.5 & 242 \\
    192 & 448 & 56\% & 43.8 & 228 & 41.3 & 288 \\
    224 & 352 & 66\% & 41.5 & 267 & 40.8 & 354 \\
    256 & 256 & 75\% & 39.9 & 424 & - & - \\
    \bottomrule
  \end{tabular}}
  \caption{\textbf{Content-aware token sharing.} The mean IoU and throughput of the Segmenter model~\cite{strudel_segmenter_2021} with ViT-S/16 backbone~\cite{dosovitskiy_image_2021} with our CTS and a random sharing approach for different token sharing settings. Results on the ADE20K val set~\cite{zhou_scene_2017}.}
  \label{tab:ade20k_grouping_schedules}
\vspace{-10pt}
\end{table}

\subsection{Ablations}

\textbf{Token sharing policy generation.} 
To analyze the role of the token sharing policy generation method $p$ in CTS, we apply and compare different methods.
These methods can be categorized into two groups: (1) using our policy network (PolicyNet) to directly predict the class-agnostic CTS policy, and (2) using semantic segmentation predictions to determine the CTS policy, \ie, following the procedure in ~\Cref{fig:grouping_examples} but with predicted semantic segmentation. The results are provided in \Cref{tab:ablations_policy}.
First of all, these results show that turning the problem into a class-agnostic classification task, instead of using the trivial solution of solving the full class-wise segmentation task, is necessary to achieve the best results. Moreover, this classification task allows us to solve it with a lightweight PolicyNet, which allows for a much higher throughput than conventional segmentation networks (162 vs.~60). Finally, we see that the performance can be boosted by having a larger PolicyNet, but at the expense of efficiency.

\begin{table}[t]
  \centering
  \setlength{\tabcolsep}{4pt}
  \adjustbox{width=\linewidth}{
  \begin{tabular}{@{}llccc@{}}
    \toprule
    \multirow{2}{*}{Category} & \multirow{2}{*}{Policy method} & Precision & Throughput & mIoU  \\
     & &  (policy) & (im/sec) &  (seg.)  \\
    \midrule
    \multirow{3}{*}{PolicyNet} & PolicyNet (default) & 80.3 & 162 & 45.1 \\
    & PolicyNet (large) & 79.3 & 148 & 45.3 \\
    & PolicyNet (small) & 73.9 & 186 & 44.0 \\
    \midrule
    \multirow{3}{*}{Seg. pred.} 
    & MobileNetV2 + FCN~\cite{sandler2018mobilenetv2, shelhamer_fully_2017} & 70.7 & 125 & 44.3\\
    & DeepLabv3+ (RN-101) ~\cite{chen_encoder-decoder_2018,he_deep_2016} & 74.2 & 60 & 44.8 \\
    & \textit{Ground-truth seg.} & \textit{96.1} & \textit{189} & \textit{47.2} \\
    \midrule
     & Random & 52.1 & 191 &  42.9 \\
    & None (baseline) & - & 122 & 45.0\\
    \bottomrule
  \end{tabular}
  }
  \caption{\textbf{Different policy methods.} Ablation of policy generation methods applied to Segmenter with ViT-S/16, on the ADE20K val set~\cite{zhou_scene_2017}. The token reduction schedule is set to 30\%.}
  \label{tab:ablations_policy}
\end{table}

\textbf{Different backbones.} The CTS approach is compatible with any ViT-based backbone that purely applies global self-attention. In \Cref{tab:ade20k_main}, we apply CTS to three backbones of similar sizes, which have different settings for weight initialization~\cite{bao2022beit}, distillation~\cite{pmlr-v139-touvron21a} and positional embeddings~\cite{bao2022beit}. For all three backbones, our CTS approach consistently results in an increase in throughput by reducing the number of tokens, without any loss in segmentation performance. This shows the general applicability of CTS.

\begin{table}[t]
  \centering
  \adjustbox{width=0.9\linewidth}{
  \begin{tabular}{@{}clccc@{}}
    \toprule
    CTS & Backbone & Token reduction & Im/sec & mIoU  \\
    \midrule
    - & \multirow{2}{*}{ViT-B/16~\cite{dosovitskiy_image_2021}}  & 0\% & 47 & 48.5 \\
    \cmark &   & 30\% & 69 & 48.7 \\
    \midrule
    - & \multirow{2}{*}{DeiT-B/16$^\dagger$~\cite{pmlr-v139-touvron21a}}  & 0\% & 48 & 48.0 \\
    \cmark &   & 30\% & 70 & 47.9 \\
    \midrule
    - & \multirow{2}{*}{BEiT-B/16~\cite{bao2022beit}} & 0\% & 36 & 50.3 \\
    \cmark &   & 30\% &  63 & 50.4 \\
    \bottomrule
  \end{tabular}}
  \caption{\textbf{Different backbones.} The mean IoU and throughput of the Segmenter model~\cite{strudel_segmenter_2021} with our CTS method, for different backbones. Results on the ADE20K val set~\cite{zhou_scene_2017}. $^\dagger$Distilled.}
  \label{tab:ade20k_main}
\end{table}

\textbf{Different decoders.} To further show the general applicability of CTS, we also show that it is compatible with multiple decoders. We apply CTS to three different types of decoders to generate the semantic predictions: 1)~\textit{Linear}: a vanilla linear projection, 2)~\textit{Mask Transformer}:  a transformer-based decoder proposed in Segmenter~\cite{strudel_segmenter_2021} and deployed in our main experiments, and 3)~\textit{UPerNet}: a commonly-used CNN-based decoder~\cite{xiao2018unified}. Again, the results in Table~\ref{tab:ade20k_multi_dec} show that CTS considerably improves the throughput, \ie, by up to 47\%, while achieving a similar segmentation quality.

\begin{table}[t]
  \centering
  \adjustbox{width=\linewidth}{
  \begin{tabular}{@{}clccc@{}}
    \toprule
    CTS & Decoder & Token reduction & Im/sec & mIoU  \\
    \midrule
    - & \multirow{2}{*}{Linear} & 0\% & 57 & 48.1 \\
    \cmark &  & 30\% & 82 & 47.9 \\
    \midrule
    - & \multirow{2}{*}{Mask Transformer \cite{strudel_segmenter_2021}} & 0\% & 47 & 48.5 \\
    \cmark &  & 30\% & 69 & 48.7 \\
    \midrule
    - & \multirow{2}{*}{UPerNet \cite{xiao2018unified}} & 0\% & 21 & 48.1 \\
    \cmark &  & 30\% & 28 & 47.7 \\
    \bottomrule
  \end{tabular}}
  \caption{\textbf{Different decoders.} Our approach is compatible with multiple decoders, including transformer-based \cite{strudel_segmenter_2021} or CNN-based \cite{xiao2018unified} ones. Each network is trained with the ViT-B/16 backbone. Results on the ADE20K val set~\cite{zhou_scene_2017}.}
  \label{tab:ade20k_multi_dec}
\end{table}

\subsection{Application to state-of-the-art}

We apply CTS to state-of-the-art method BEiT-L + UPerNet~\cite{bao2022beit,xiao2018unified}, and report the results in~\Cref{tab:ade20K_sota}. For reference, we also show the results of other state-of-the-art networks with similar backbones and pre-training. The results show that our CTS significantly improves the throughput, by nearly 64\%, while achieving a similar segmentation quality. This shows that CTS is also effective when applied to state-of-the-art methods, that it is compatible with large-scale self-supervised pre-training methods~\cite{bao2022beit} that have shown to significantly boost the performance for downstream tasks, and that it is competitive with top-performing ViT-based segmentation networks.

\begin{table}[t]
  \centering
  \adjustbox{width=\linewidth}{
  \begin{tabular}{@{}lcccc@{}}
    \toprule
    \multirow{2}{*}{Method} & \multirow{2}{*}{Pre-training} & Token & \multirow{2}{*}{Im/sec}& \multirow{2}{*}{mIoU (ms)}  \\
    & & reduction & & \\
    \midrule
    BEiT-L* + UPerNet & BEiT + IN22k & 0\% & 3.3 & 56.8 \\
    BEiT-L* + UPerNet + CTS (ours)  & BEiT + IN22k & 30\% & 5.4 & 56.6 \\
    \midrule
    \color{darkgray} Swin-L + UperNet~\cite{dong2022cswin,xiao2018unified} & \color{darkgray} IN22k  & \color{darkgray} n/a &   & \color{darkgray} 53.5 \\
    \color{darkgray} ViT-Adapter-L + UperNet~\cite{chen2022vitadapter,xiao2018unified} & \color{darkgray} IN22k & \color{darkgray} n/a &  & \color{darkgray} 54.4 \\
    \color{darkgray} SwinV2-L + UPerNet~\cite{liu2022swinv2,cheng2022mask2former} & \color{darkgray} IN22k & \color{darkgray} n/a &  & \color{darkgray} 55.9 \\
    \color{darkgray} Swin-L + Mask2Former~\cite{liu_swin_2021,cheng2022mask2former} & \color{darkgray} IN22k & \color{darkgray} n/a &  & \color{darkgray} 57.3 \\
    \color{darkgray} BEiT-L + UPerNet~\cite{bao2022beit,xiao2018unified} & \color{darkgray} BEiT + \color{darkgray} IN22k & \color{darkgray} n/a & & \color{darkgray} 57.0 \\
    \color{darkgray} ViT-Adapter-L + UperNet~\cite{chen2022vitadapter,xiao2018unified} & \color{darkgray} BEiT + \color{darkgray} IN22k & \color{darkgray} n/a & & \color{darkgray} 58.4 \\
    \color{darkgray} ViT-Adapter-L + Mask2Former~\cite{chen2022vitadapter} & \color{darkgray} BEiT + IN22k & \color{darkgray} n/a &  & \color{darkgray} 59.0 \\
    \bottomrule
  \end{tabular}}
  \caption{\textbf{Application to state-of-the-art.} We apply CTS to state-of-the-art method BEiT-L + UPerNet~\cite{bao2022beit,xiao2018unified} and report other top-performing methods for reference. Results on the ADE20K~\cite{zhou_scene_2017} dataset, with multi-scale testing. *Our implementation~\cite{mmseg2020}.}
  \label{tab:ade20K_sota}
\vspace{-10pt}
\end{table}

\begin{table}[t]
  \centering
  \adjustbox{width=\linewidth}{
  \begin{tabular}{@{}llccc@{}}
    \toprule
    Method & Backbone & Token reduction & Im/sec & mIoU  \\
    \midrule
    \multicolumn{5}{c}{\textit{Pascal Context validation}} \\
    \midrule
    Segmenter & ViT-S/16 & 0\% &  157 & 53.0 \\
    Segmenter + CTS (rand.)  & ViT-S/16 & 30\% & 246 & 51.3 \\
    Segmenter + CTS (ours)  & ViT-S/16 & 30\% & 203 & 52.9 \\
    \midrule
    Segmenter & ViT-L/16 & 0\% &  21 & 57.7 \\
    Segmenter + CTS (ours) & ViT-L/16 & 30\%  & 31 & 57.6 \\
    \midrule
    \multicolumn{5}{c}{\textit{Cityscapes val}} \\
    \midrule
    Segmenter & ViT-S/16 & 0\%  & 38 & 76.5 \\
    Segmenter + CTS (rand.) & ViT-S/16 & 44\%  & 93 & 72.6 \\
    Segmenter + CTS (ours) & ViT-S/16 & 44\%  & 78 & 76.5 \\
    \midrule
    Segmenter & ViT-L/16 & 0\% & 6.0 & 79.1 \\
    Segmenter + CTS (ours)  & ViT-L/16 & 44\% & 12.6 & 79.5 \\
    \bottomrule
  \end{tabular}
  }
  \caption{\textbf{Other datasets.} CTS with Segmenter~\cite{strudel_segmenter_2021} on the Pascal Context \cite{mottaghi_role_2014} and Cityscapes~\cite{cordts_cityscapes_2016} datasets.}
  \label{tab:pascal_main}
\vspace{-5pt}
\end{table}

\begin{table*}[t]
\adjustbox{width=\linewidth}{
\subfloat[\textbf{Consistency in predictions for shared tokens.} We force all pixels in superpatches to have the same semantic class prediction if they share a token, and evaluate the mean IoU.\label{tab:force_one_cls}]{
\tablestyle{4.8pt}{1.05}\begin{tabular}{@{}cccr@{}}
    \toprule
    \multirow{2}{*}{Token reduction} & \multicolumn{3}{c}{mIoU}  \\
    \cmidrule{2-4}
    & default & one class & diff.  \\
    \midrule
    12\% & 44.9 & 44.9 & 0.0 \\
    30\% & 45.1 & 45.1 & 0.0\\
    56\% & 43.8 & 43.4 & -0.4\\
    66\% & 41.5 & 40.8 & -0.7\\
        \bottomrule
\vspace{10pt}
  \end{tabular}}
  \hspace{3mm}
\subfloat[\textbf{Dynamic token sharing.} Results of our approach when images are assigned to models with different token sharing settings, depending on the predicted policy.
  \label{tab:ensemble}]{
\tablestyle{2.5pt}{1.05}  \begin{tabular}{@{}ccc@{}}
    \toprule
    Token reduction & Processed images & mIoU  \\
    \midrule
    0\% & 2 & \textit{45.0} \\
    12\% & 87 & \textit{44.9} \\
    23\% & 191 & \textit{45.0} \\
    30\% & 398 & \textit{45.1} \\
    38\% & 427 & \textit{44.7} \\
    46\% & 895 & \textit{44.4} \\
    \midrule
    Combined & 2000 & \textbf{45.8}\\ 
    \bottomrule
  \end{tabular}}\hspace{3mm}

\subfloat[\textbf{CTS on video data.} Experiments with video inputs on the Cityscapes val set~\cite{cordts_cityscapes_2016}. The token reduction setting is 44\%. \label{tab:video_cityscapes}]{
\tablestyle{2.2pt}{1.05}
\begin{tabular}{clcc}
    \toprule
    Using & \multirow{2}{*}{Policy} & Precision & mIoU  \\
    prev.~frame& & (policy) & (seg.)\\
    \midrule
    \cmark & Segmentation pred. & 85.9 & 75.5 \\
    \cmark & PolicyNet & 94.9 & 76.2\\
    \midrule
    \xmark & PolicyNet (default) & 96.8 & 76.5   \\
    \xmark & Random & 69.5 &  72.6 \\
    \xmark & None (baseline) & - & 76.5   \\
    \xmark & \textit{Ground truth} & \textit{99.6} & \textit{78.0} \\
    \bottomrule
  \end{tabular}
}\vspace{-1mm}
}
\vspace{-3mm}
\caption{
\textbf{Additional analyses.} Networks trained with the ViT-S/16 backbone. All results on ADE20K val~\cite{zhou_scene_2017} unless indicated otherwise.
\label{tab:ablations}}
\vspace{-10pt}
\end{table*}

\subsection{Other datasets}
To show that CTS is also effective on other datasets, we evaluate it on Pascal Context~\cite{mottaghi_role_2014} and Cityscapes~\cite{cordts_cityscapes_2016}. We apply CTS to Segmenter, report the results in \Cref{tab:pascal_main}, and compare to random token sharing for reference.
From the results, it is clear that CTS consistently achieves a mean IoU similar to the standard Segmenter, while improving the throughput by up to 48\% for Pascal Context and 110\% for Cityscapes. The higher efficiency boost for Cityscapes is caused by a higher token reduction, which is possible because this dataset contains images in which more superpatches contain just a single class, as depicted in \Cref{fig:patch_histograms}, which means that more patches can share a token without diminishing the segmentation performance.

\subsection{Additional analyses}

\textbf{Consistency in predictions for shared tokens.} %
With our policy network, we aim to let superpatches share a token if they contain a single semantic class, and in \Cref{tab:ablations_policy} we have shown that the network mostly correctly predicts this. However, it is still unclear if the segmentation network subsequently also predicts that these superpatches contain a single semantic class. To check this, we enforce that each pixel in a superpatch that shares a token gets the same class prediction, and evaluate the impact on the mean IoU. We achieve this single-class prediction with a majority vote in each superpatch that shares a token. The results, reported in \Cref{tab:force_one_cls}, show that for token reductions up to 30\%, the mean IoU remains constant. This indicates that there is little to no difference in the predictions, meaning that the network already makes consistent predictions within superpatches that share tokens. This further shows that our policy network is effective, and that the segmentation network automatically learns that these superpatches contain only one class, which is desired behavior.

\textbf{Dynamic token sharing. } The proposed CTS method uses a fixed number of $\text{S}$ shared tokens for all the images in a dataset. However, images typically have different semantic complexities, even in the same dataset (see \Cref{fig:patch_histograms}). To evaluate whether it is desirable to let $\text{S}$ depend on the complexity of the image, we conduct an experiment where we first apply our policy network to each image, threshold its prediction, and identify how many superpatches $\text{S}^*$ can share a token according to this prediction. Subsequently, we pick the highest token sharing setting S for which $\text{S}<\text{S}^*$, let the S highest-scoring superpatches share a token, and feed the resulting tokens through a segmentation network trained with this setting $\text{S}$. Finally, we collect all segmentation predictions and calculate the mean IoU.

\Cref{tab:ensemble} presents the results for this experiment, together with the number of images assigned to each model. Interestingly, we find that the resulting mean IoU is 45.8, which is 0.7 points higher than the best individual model. This shows that, by dynamically determining how many tokens should be shared in each image, the segmentation quality can be even higher, because we do not force superpatches to share a token if the policy network assigns them a low score. This means that a future single-network, variable-token approach could optimize the accuracy and efficiency even further. Because the number of shared tokens in such a method would vary, there would be no guaranteed efficiency boost, introducing a new accuracy-efficiency trade-off.

\textbf{CTS on video data.} In the situation where there is a continual stream of images, \ie, with video data, there is a large degree of consistency between subsequent frames. If this consistency is high enough, the predictions for a previous frame could be used to generate a policy for a current frame, removing the need for a separate policy network, which could improve the efficiency.
We evaluate this for two settings: (1) we use the previous segmentation predictions to generate a policy according to the procedure in \Cref{fig:grouping_examples}, and (2) we run our policy network on the previous frame and use its output as the policy. The results, presented in \Cref{tab:video_cityscapes}, show that using information from previous frames results in a segmentation quality only slightly below the original CTS approach, especially for the PolicyNet, \ie, -0.3. Such an approach can be more efficient than the original policy network if the policy network is executed in parallel with the segmentation network, and it could be interesting for future research. Again, this experiment also shows the strength of our policy network, as it outperforms the trivial solution of using the previous semantic segmentation output as a policy.

\section{Discussion}
The experimental results demonstrate the positive impact of token sharing on the efficiency of ViT-based semantic segmentation networks. We show that the number of tokens can be reduced by 30\% to 44\% without compromising segmentation quality, depending on the dataset. Key to maintaining segmentation quality is selecting the correct superpatches for token sharing, \ie, the superpatches containing a single semantic class. Our CTS approach achieves this with an efficient policy network that is conveniently trained in a supervised manner using the already available ground-truth data. Especially for large ViT models applied to high-resolution images, the computational overhead that our policy network introduces is marginal, and a significant improvement in throughput is achieved.

The strength of our CTS framework is its conceptual simplicity, which allows it to be very efficient and generally applicable, making it compatible with many vision transformers, including those using advanced pre-training. Such pre-training methods have shown to be key to obtaining state-of-the-art results~\cite{bao2022beit} and our CTS enables such current and future methods to be significantly more efficient.

Due to its simple nature, CTS can be extended in several different ways in future work. First of all, it is conceptually compatible with alternative efficient approaches that combine global and local self-attention, such as Swin~\cite{liu_swin_2021}, and these methods have complementary advantages. Second, our token sharing approach is also compatible with other computation-intensive per-pixel tasks, such as optical flow and depth estimation. CTS can be applied to improve the efficiency on those tasks, but would require task-specific policy models. With this work, we lay the foundation for such follow-up research on efficient transformer-based architectures for per-pixel computer vision tasks.

\label{sec:conclusion}

{\small \paragraph{Acknowledgements}
This work is supported by Eindhoven Engine, NXP Semiconductors and Brainport Eindhoven. This work made use of the Dutch national e-infrastructure with the support of the SURF Cooperative using grant no.~EINF-3836, which is financed by the Dutch Research Council (NWO).}

{\small
\bibliographystyle{ieee_fullname}
\bibliography{bibliography.bib}
}

\clearpage

\appendix 

\section*{Appendix}

This appendix contains the following supplementary material:

\begin{itemize}
    \item More extensive implementation details (\Cref{sec:app:impl_details}).
    \item Detailed results for the dynamic token sharing experiment (\Cref{sec:app:dynamic_token_sharing}). 
    \item Additional qualitative results, showing predictions by both the policy network and the complete semantic segmentation network (\Cref{sec:app:qual_results}).
\end{itemize}

\noindent The code for Content-aware Token Sharing (CTS) is available through \url{https://tue-mps.github.io/CTS}.

\section{Implementation details}
\label{sec:app:impl_details}
Below, we provide the implementation details for the networks used in our experiments. Note that, for our main experiments, we conducted each experiment 5 times, and report the median mIoU. 

\subsection{Policy network} Our default Content-aware Token Sharing (CTS) policy network, used to identify what superpatches should share a token, is based on EfficientNet~\cite{tan_efficientnet_2019}. Specifically, we use EfficientNet-Lite0, a light version of EfficientNet-B0, pre-trained on ImageNet-1K~\cite{russakovsky_imagenet_2015}, using the code and pre-trained model weights from the \textit{timm} repository~\cite{rw2019timm}. To be able to make a prediction per superpatch, we first output the features before the final classification head. By design of the network, these features have resolution ${H_p} \times {W_p}$ that is 32 times smaller than the resolution of the input images. As all segmentation networks used in this work use a patch size of $16\times16$ pixels, a superpatch is $32\times32$ pixels. Therefore, each of the $H_p \cdot W_p$ features represent one superpatch. To get a policy prediction per superpatch, we add one final $1\times1$ convolutional layer that maps the features to two classes, \ie, (1) the superpatch contains a single class and (2) the superpatch contains multiple classes. We train the policy network separately for ADE20K~\cite{zhou_scene_2017}, Pascal Context~\cite{mottaghi_role_2014} and Cityscapes~\cite{cordts_cityscapes_2016}. The policy network is always trained for 50k iterations with batches of 8 image crops, a learning rate of $5\times{10}^{-5}$ and a weight decay of ${10}^{-3}$, using the AdamW optimizer~\cite{loshchilov2018decoupled}.

\paragraph{Large and small policy networks.}
In Table 2 of the main manuscript, we also provide results for a large and small policy network. The large policy network is EfficientNet-B1~\cite{tan_efficientnet_2019}, pre-trained on ImageNet-1K~\cite{russakovsky_imagenet_2015}, using the code and pre-trained weights from the \textit{timm} repository~\cite{rw2019timm}. All hyperparameters are the same as for the default policy network. The small policy network is a custom neural network that has seven $3\times3$ convolutional layers, with 128 output channels per convolution, and a final $1\times1$ convolution to output a prediction for two classes. Strided convolutions are used to reduce the resolution by a factor of 32. This network is not pre-trained, and we train it with a learning rate of $10^{-3}$.

\subsection{Segmenter} 
\label{sec:impl_details:segmenter}
In our main experiments, we apply our CTS to a transformer-based semantic segmentation network, Segmenter~\cite{strudel_segmenter_2021}. For these experiments, we implement CTS in the official, publicly available code of Segmenter, and we use the original training settings. That means that we use a batch size of 8 for experiments with ADE20K and Cityscapes and a batch size of 16 for Pascal Context, use the Stochastic Gradient Descent (SGD) optimizer with a momentum of 0.9, a weight decay of 0.0, and a polynomial decay learning rate schedule. For ADE20K, we train for 64 epochs with an initial learning rate of 0.001, resize the shortest side of images to 512 pixels, and take random crops of $512\times512$ pixels. For the \textit{ViT-L (640)} setting in Figure 5 of the main manuscript, we resize the shortest side of images to 640 pixels, and take random crops of $640\times640$ pixels. For Pascal Context, we train for 256 epochs with an initial learning rate of 0.001, resize the shortest side of images to 520 pixels, and take random crops of $480\times480$ pixels. For Cityscapes, we train for 216 epochs with an initial learning rate of 0.01, and take random crops of $768\times768$ pixels. 

\paragraph{DeiT and BEiT.} 
For our experiment with different backbones, we use DeiT-B and BEiT-B in addition to the standard ViT backbones. For DeiT-B~\cite{pmlr-v139-touvron21a}, we follow the settings from Segmenter, and initialize the backbone with distilled weights from the \textit{timm} repository~\cite{rw2019timm}, pre-trained on ImageNet-1K. In the experiments with the BEiT-B backbone~\cite{bao2022beit}, we also use the code and pre-trained weights from \textit{timm}, but we make minor changes for compatibility: (1) We use the AdamW optimizer with an initial learning rate of $10^{-5}$ and weight decay of $10^{-3}$ because the SGD optimizer made training unstable. (2) Because BEiT works with relative position embeddings between all tokens, these have to be shared too, when a superpatch shares a token. We empirically find that picking the relative position to one of the patches in a superpatch works just as well as taking the mean of the relative positions to each of them, while being significantly faster.

\begin{table*}[t]%
    \centering
    \begin{tabular}{@{}ccccccccc@{}}
    \toprule
    \multirow{2}[2]{*}{Token reduction} & mIoU & \multicolumn{7}{c}{Processed images with different thresholds $\tau$} \\
    \cmidrule{3-9}
    & (indiv. models) & $\tau=0.35$ & $\tau=0.4$ & $\tau=0.45$ & $\tau=0.5$ & $\tau=0.55$ & $\tau=0.6$ & $\tau=0.65$ \\
    \midrule
    0\%  & 45.0 & 1 & 2 & 3 & 8 & 12 & 22 & 32\\
    12\% & 44.9 & 77 & 87 & 105 & 134 & 163 & 196 & 232\\
    23\% & 45.0 & 150 & 191 & 219 & 244 & 270 & 289 & 316\\
    30\% & 45.1 & 366 & 398 & 436 & 455 & 460 & 465 & 463\\
    38\% & 44.7 & 446 & 427 & 410 & 397 & 394 & 392 & 386\\
    46\% & 44.4 & 960 & 895 & 827 & 762 & 701 & 636 & 571\\
    \midrule
    \multicolumn{2}{l}{Average token reduction} & 38.2\% & 37.4\% & 36.5\% & 35.5\% & 34.6\% & 33.6\% & 32.4\% \\
    \multicolumn{2}{l}{Combined mIoU} & 45.4 & \textbf{45.8} & 45.5 & 45.3 & 45.3 & 45.2 & 45.4\\ 
    \bottomrule
    \end{tabular}
    \caption{\textbf{Dynamic token sharing ablation.} Results for the dynamic token sharing experiment for different settings of threshold $\tau$. Experiments with Segmenter~\cite{strudel_segmenter_2021} and ViT-S/16~\cite{dosovitskiy_image_2021} on the ADE20K validation set~\cite{zhou_scene_2017}.}
    \label{tab:dyna_token_sharing}
\end{table*}

\subsection{UPerNet} 
\paragraph{ViT + UPerNet.}
For our experiments with the UPerNet~\cite{xiao2018unified} decoder in combination with the standard ViT backbone~\cite{dosovitskiy_image_2021,steiner2022trainvit}, we use the publicly available code and hyperparameters from the \textit{mmsegmentation} repository~\cite{mmseg2020}, and implement CTS in this code. This means that we train these networks for 160k iterations, with a batch size of 16. We first resize the shortest side of images to 512 pixels, and then take random crops of $512\times512$ pixels. We use the AdamW optimizer, with a learning rate of $6\times{10}^{-5}$ and a weight decay of 0.01.

\paragraph{BEiT + UPerNet.} When applying our CTS to state-of-the-art network BEiT-L + UPerNet~\cite{bao2022beit,xiao2018unified}, we also use the publicly available code and hyperparameters from the \textit{mmsegmentation} repository~\cite{mmseg2020}. The weights of BEiT-L are initialized using the publicly available weights from \textit{timm}~\cite{rw2019timm}. We train the networks with a batch size of 8, for 320k iterations. We first resize the shortest side of images to 640 pixels, and then take random crops of $640\times640$ pixels. The networks are optimized using AdamW, with a learning rate of $2\times{10}^{-5}$ and a weight decay of 0.05. Again, we handle the relative positions embeddings as described for BEiT-B in \Cref{sec:impl_details:segmenter}.

\section{Dynamic token sharing}
\label{sec:app:dynamic_token_sharing}

In Section 5.5 of the main manuscript, we analyze whether it is desirable to let the number of shared tokens, S, depend on the complexity of the image. To evaluate this, we feed each image through the policy network, and use its predictions to determine how many superpatches can share a token. If the predicted score for a superpatch is above a threshold $\tau$, we assume that this superpatch contains a single class, and can therefore share a token. The resulting number of superpatches that can share a token for each image is then given by $\text{S}^*$. Then, we make sure that a maximum of $\text{S}^*$ superpatches are shared per image. Specifically, from a set of models trained with different token sharing settings S, we pick the model with the highest token sharing setting S for which $\text{S}<\text{S}^*$. Subsequently, we let the S highest-scoring superpatches share a token, and feed the resulting tokens through the segmentation model trained with setting S. In \Cref{tab:dyna_token_sharing}, we provide the results for this analysis for different values of threshold $\tau$.
It can be observed that, for all evaluated thresholds, the combined mIoU is consistently higher than the highest mIoU among all the individual models. Note that this high performance is obtained by processing more images by the models with higher token reduction settings, which have a lower individual mIoU when processing all images. Moreover, the average token reduction is above the 30\% reduction setting which was found to be optimal when the number of shared tokens is fixed. This makes dynamic token sharing for semantic segmentation an interesting topic for future research.

\section{Qualitative results}
\label{sec:app:qual_results}

\begin{figure*}[t]
\centering
\includegraphics[width=0.195\linewidth]{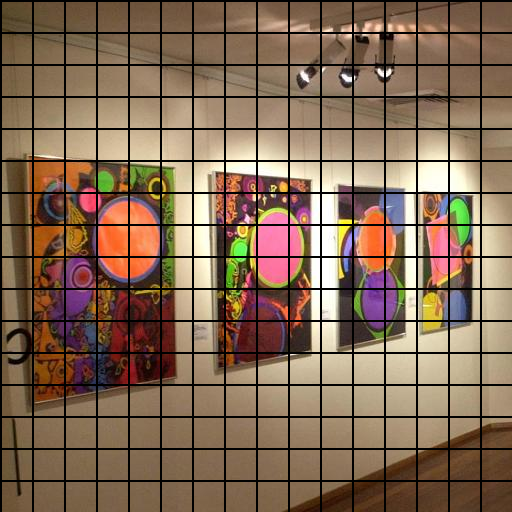}
\includegraphics[width=0.195\linewidth]{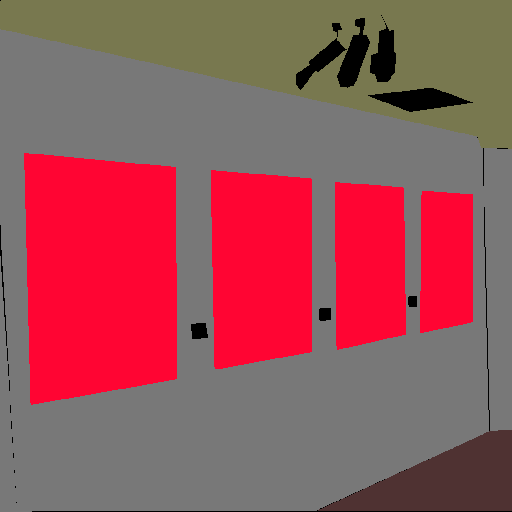}
\includegraphics[width=0.195\linewidth]{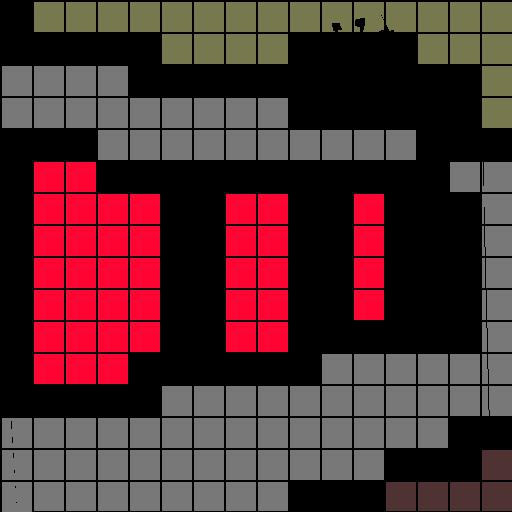}
\includegraphics[width=0.195\linewidth]{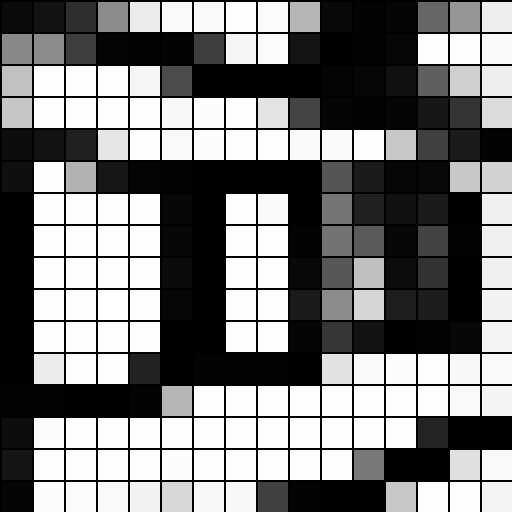}
\includegraphics[width=0.195\linewidth]{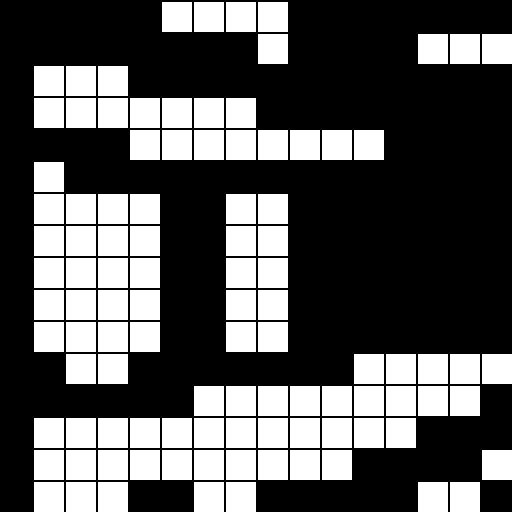} \\

\includegraphics[width=0.195\linewidth]{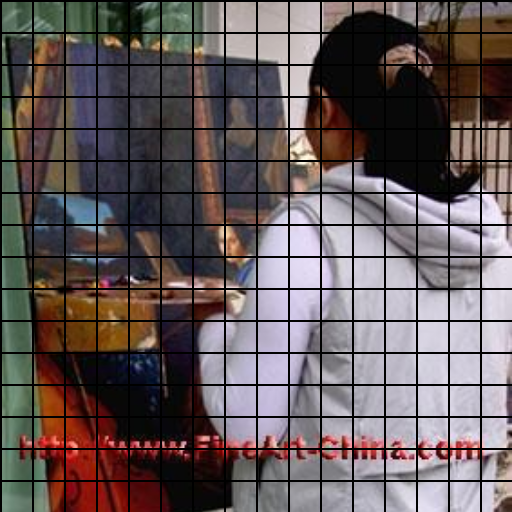}
\includegraphics[width=0.195\linewidth]{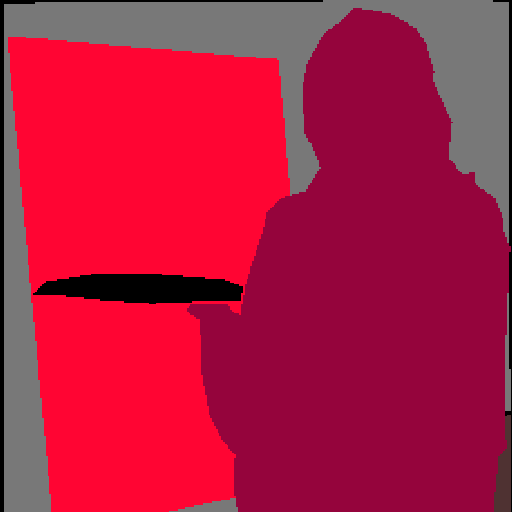}
\includegraphics[width=0.195\linewidth]{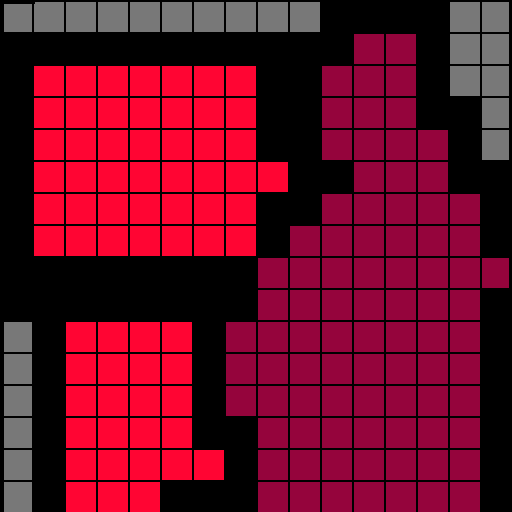}
\includegraphics[width=0.195\linewidth]{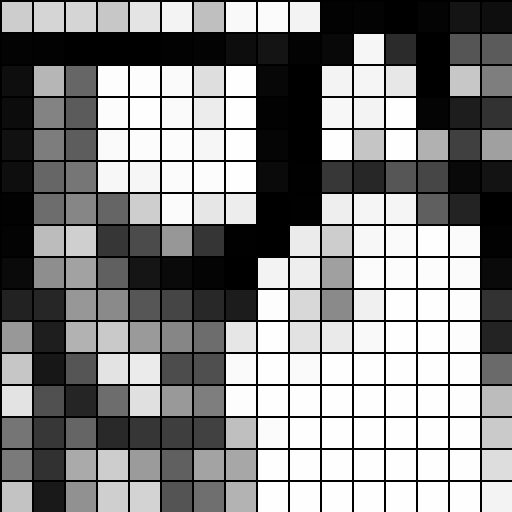}
\includegraphics[width=0.195\linewidth]{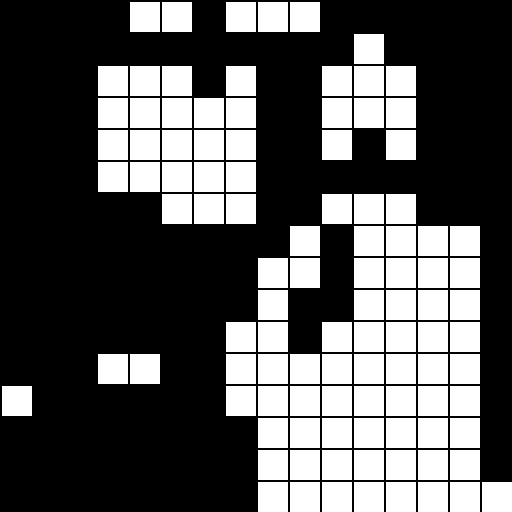} \\

\includegraphics[width=0.195\linewidth]{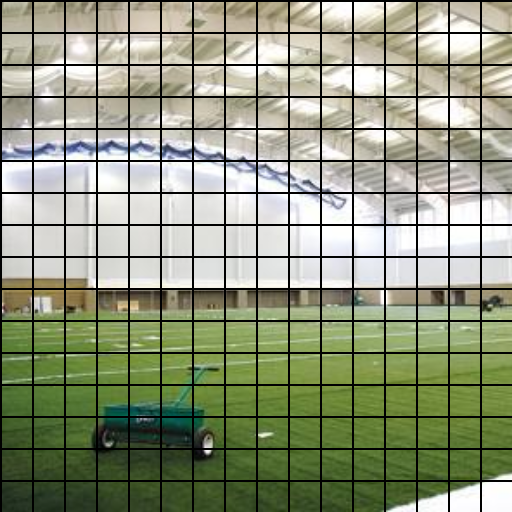}
\includegraphics[width=0.195\linewidth]{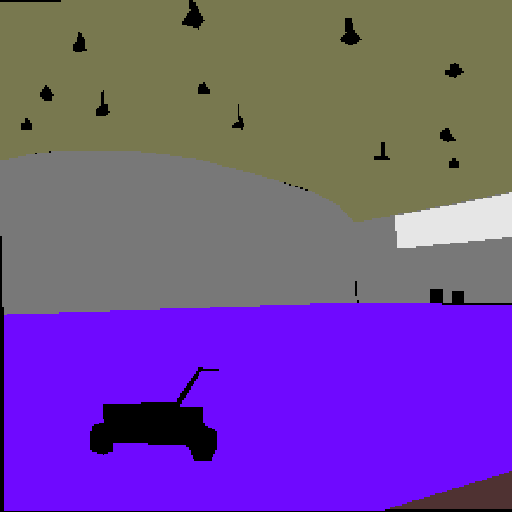}
\includegraphics[width=0.195\linewidth]{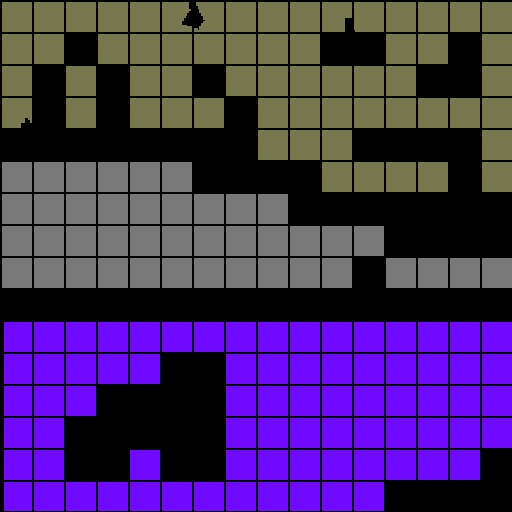}
\includegraphics[width=0.195\linewidth]{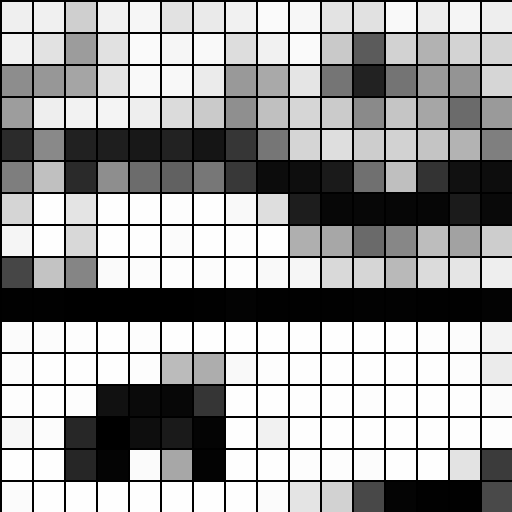}
\includegraphics[width=0.195\linewidth]{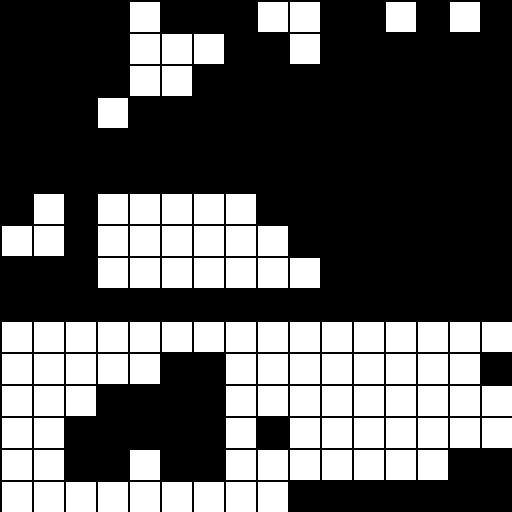} \\

\includegraphics[width=0.195\linewidth]{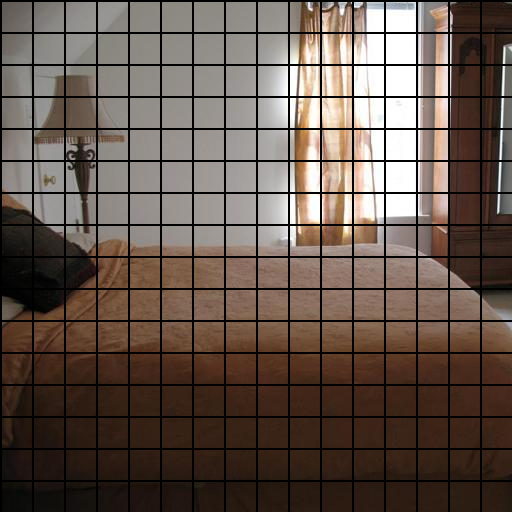}
\includegraphics[width=0.195\linewidth]{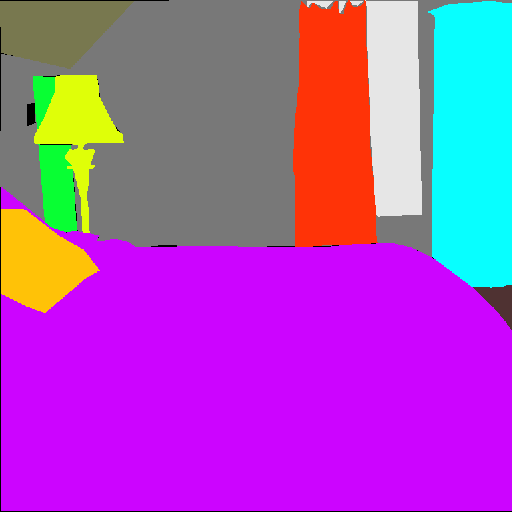}
\includegraphics[width=0.195\linewidth]{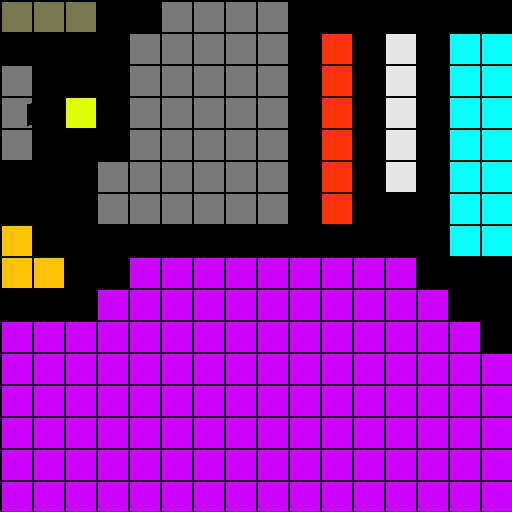}
\includegraphics[width=0.195\linewidth]{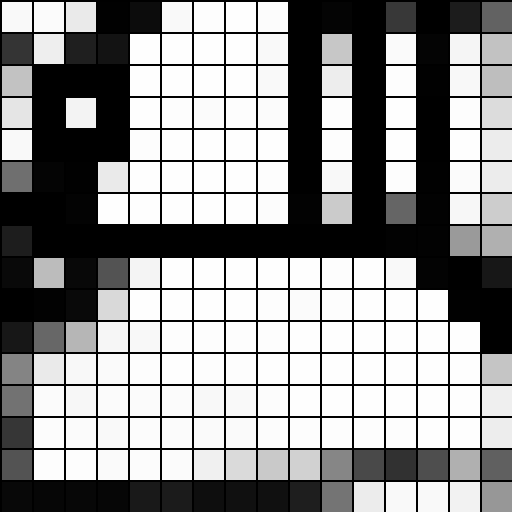}
\includegraphics[width=0.195\linewidth]{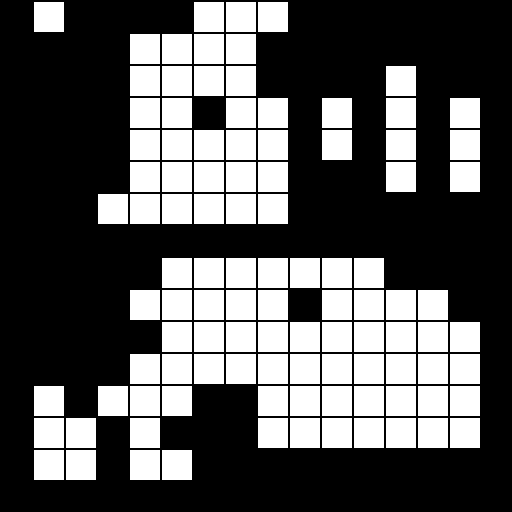} \\

\includegraphics[width=0.195\linewidth]{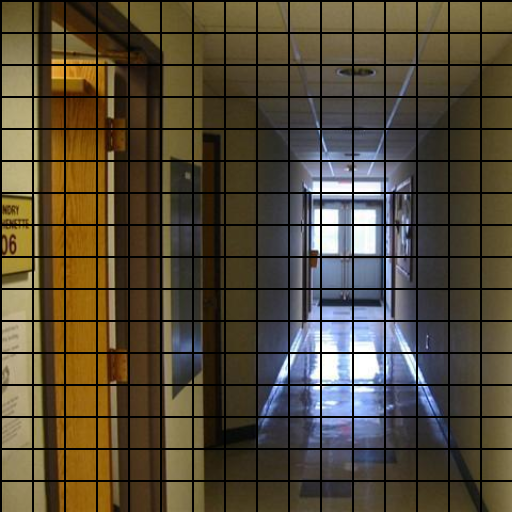}
\includegraphics[width=0.195\linewidth]{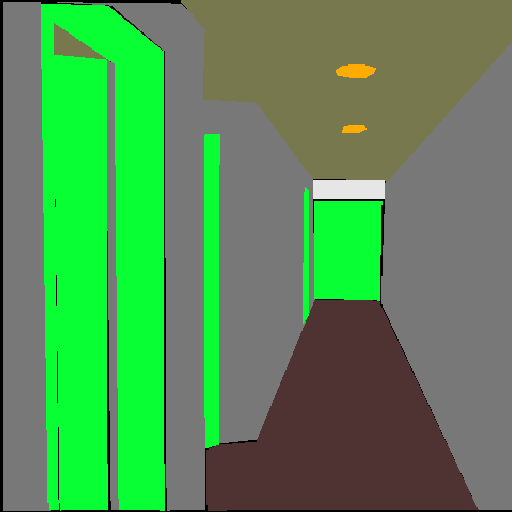}
\includegraphics[width=0.195\linewidth]{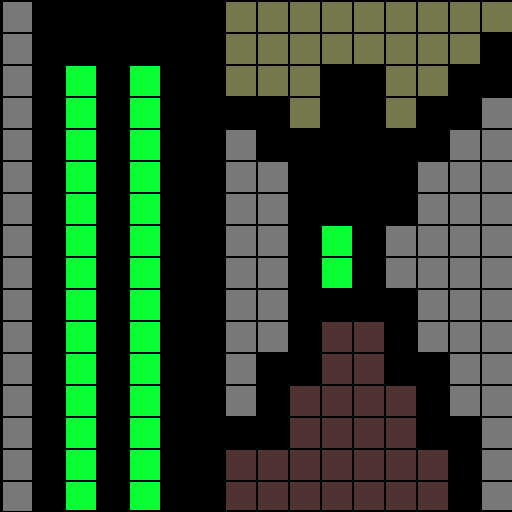}
\includegraphics[width=0.195\linewidth]{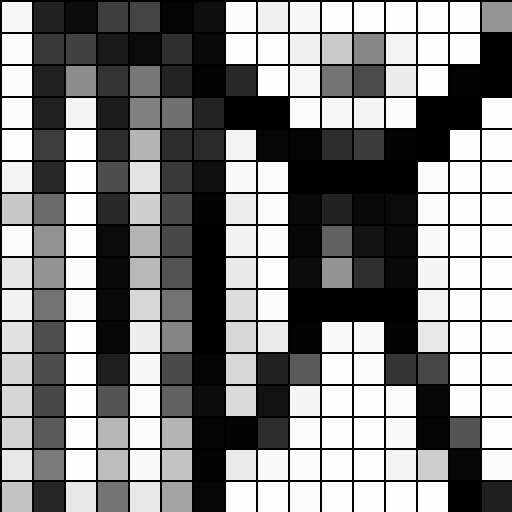}
\includegraphics[width=0.195\linewidth]{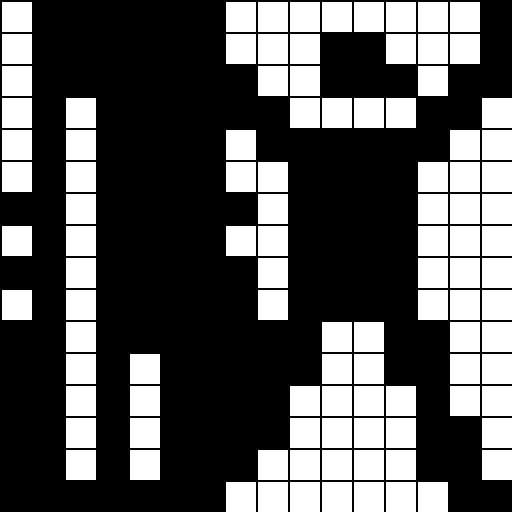} \\

\includegraphics[width=0.195\linewidth]{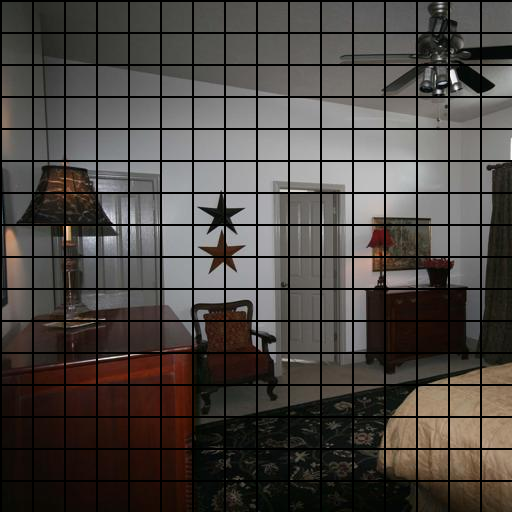}
\includegraphics[width=0.195\linewidth]{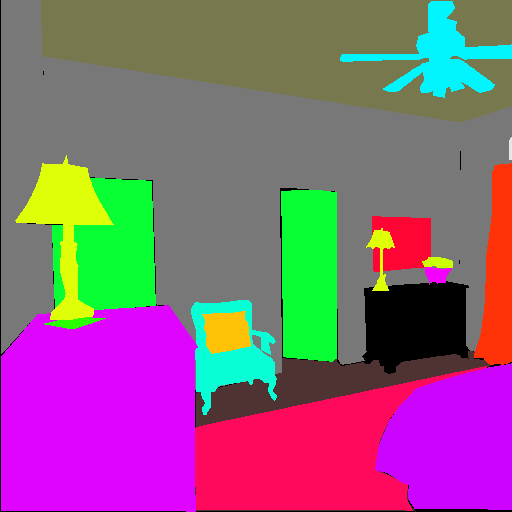}
\includegraphics[width=0.195\linewidth]{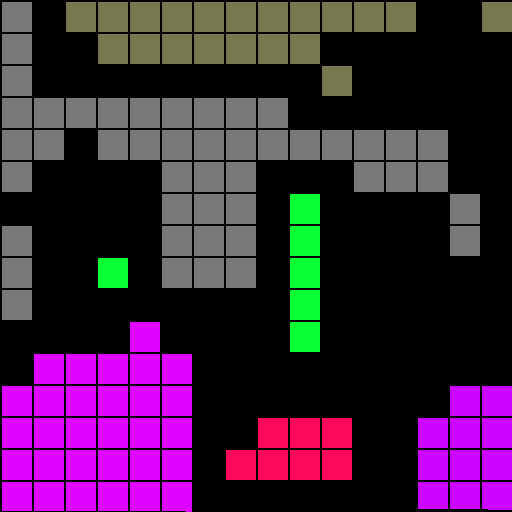}
\includegraphics[width=0.195\linewidth]{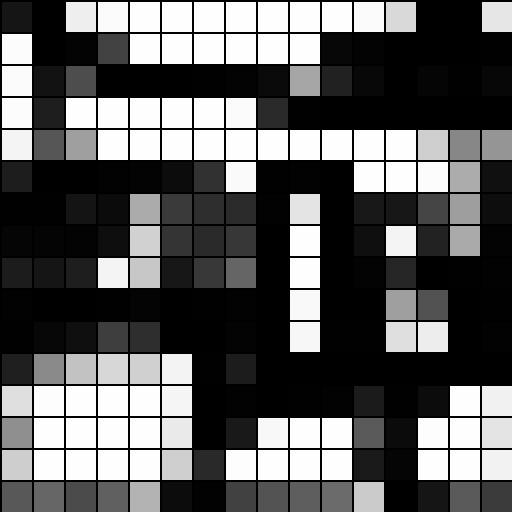}
\includegraphics[width=0.195\linewidth]{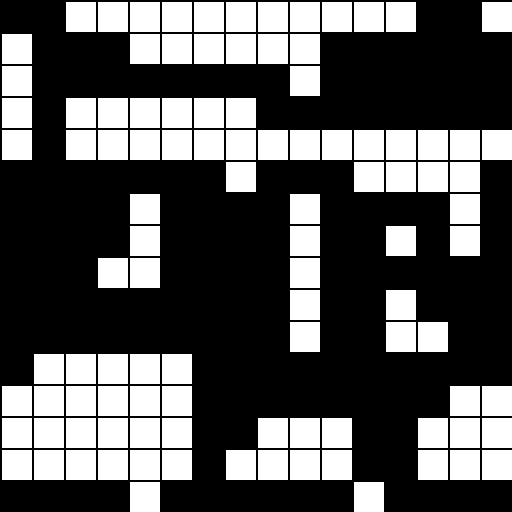} \\

 \begin{subfigure}[b]{0.195\textwidth}
     \centering
     \caption{Image with grid of superpatches}
 \end{subfigure}
 \begin{subfigure}[b]{0.195\textwidth}
     \centering
     \caption{Semantic segmentation ground truth}
 \end{subfigure}
  \begin{subfigure}[b]{0.195\textwidth}
     \centering
     \caption{Superpatches with a single class}
 \end{subfigure}
 \begin{subfigure}[b]{0.195\textwidth}
     \centering
     \caption{Policy network prediction\ \ \ \ \ \ \ \ }
 \end{subfigure}
\begin{subfigure}[b]{0.195\textwidth}
     \centering
     \caption{S highest-scoring superpatches (\ie, used policy)}
 \end{subfigure}

\caption{\textbf{Qualitative results policy network.} Predictions by our content-aware token sharing policy network, on image crops from the ADE20K validation set~\cite{zhou_scene_2017}.} 
\label{fig:policynet_qual_results_ade20k}
\end{figure*}

\begin{figure*}[t]
\centering
\includegraphics[width=0.195\linewidth]{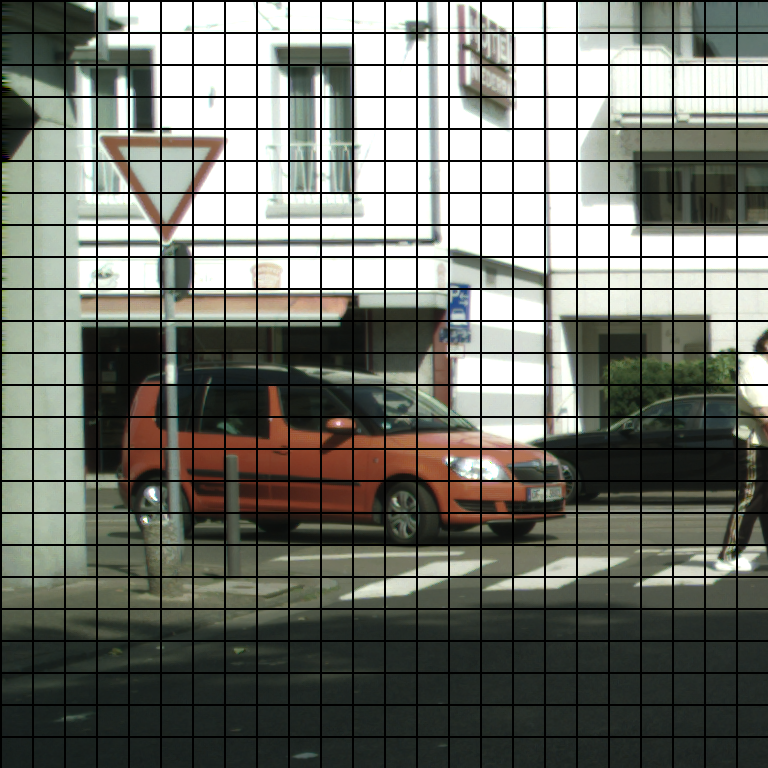}
\includegraphics[width=0.195\linewidth]{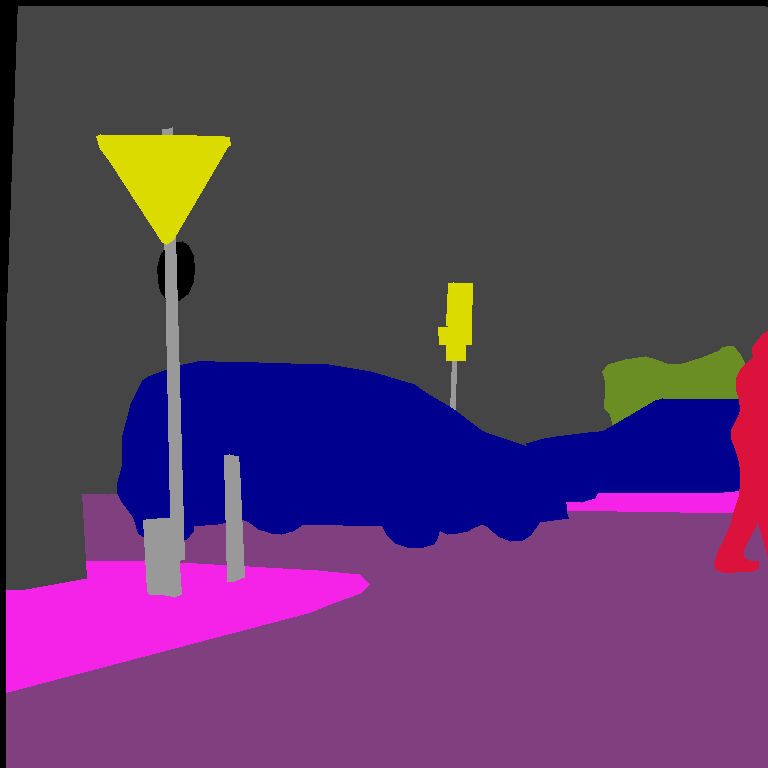}
\includegraphics[width=0.195\linewidth]{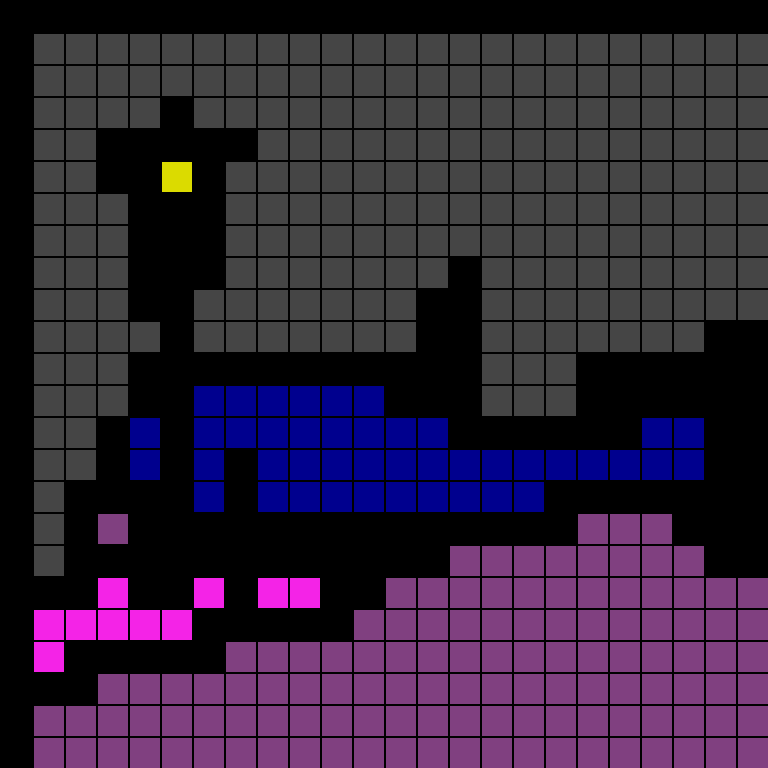}
\includegraphics[width=0.195\linewidth]{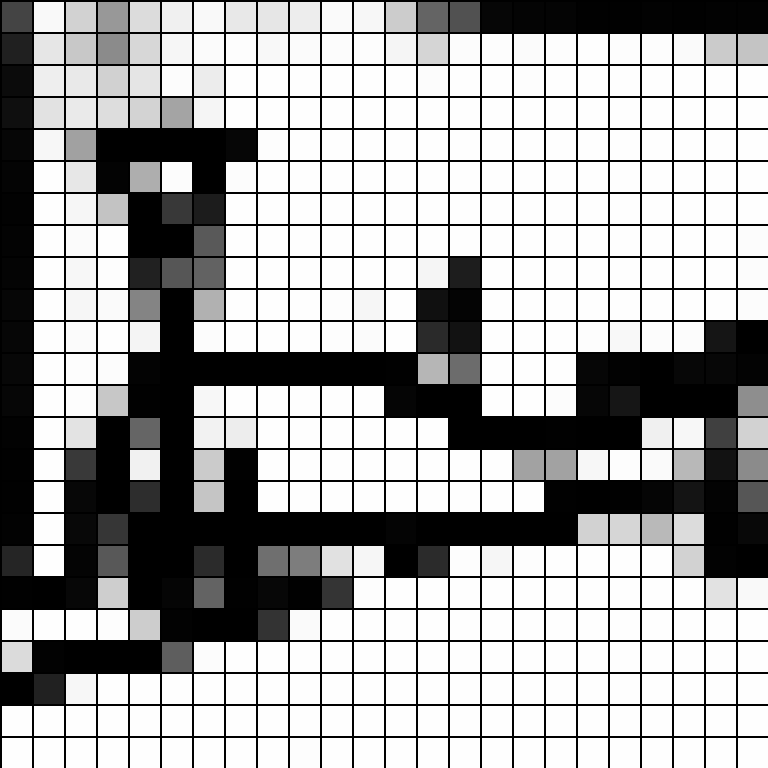}
\includegraphics[width=0.195\linewidth]{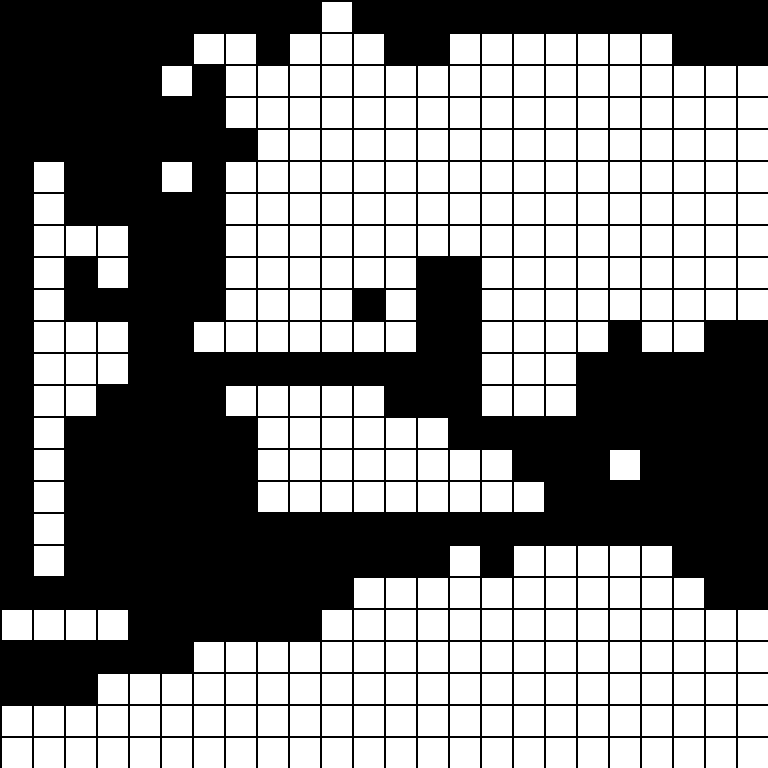} \\

\includegraphics[width=0.195\linewidth]{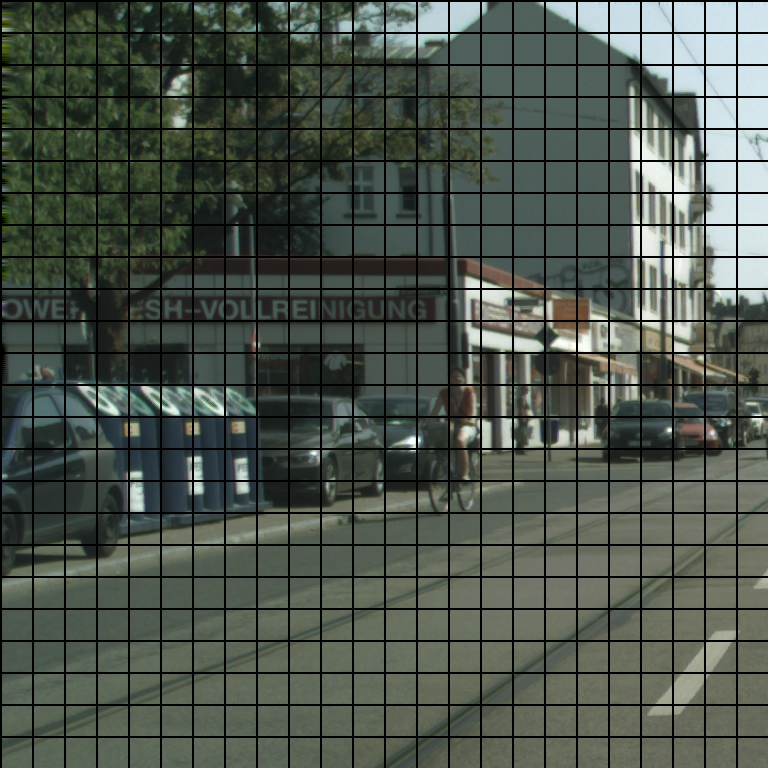}
\includegraphics[width=0.195\linewidth]{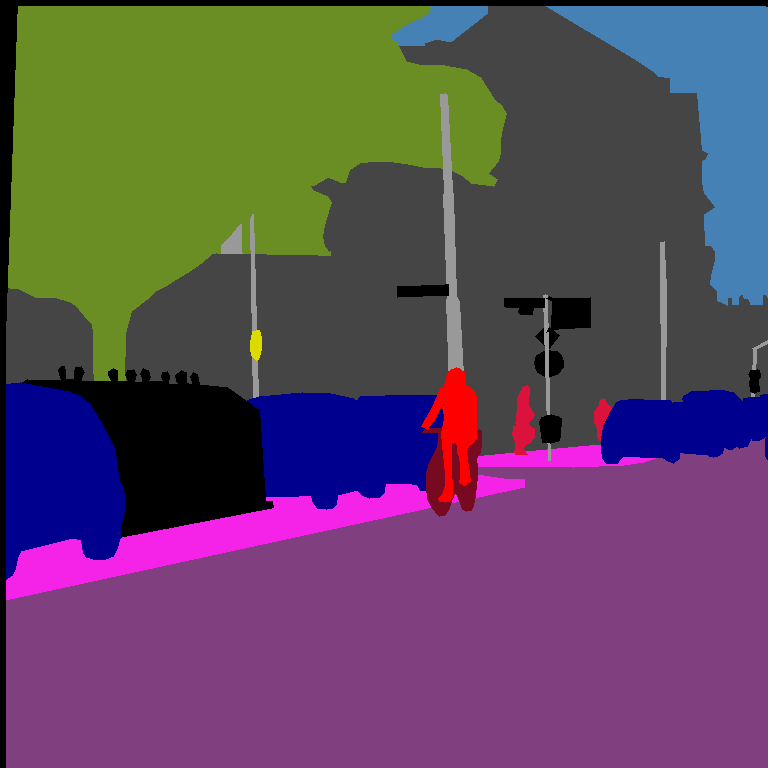}
\includegraphics[width=0.195\linewidth]{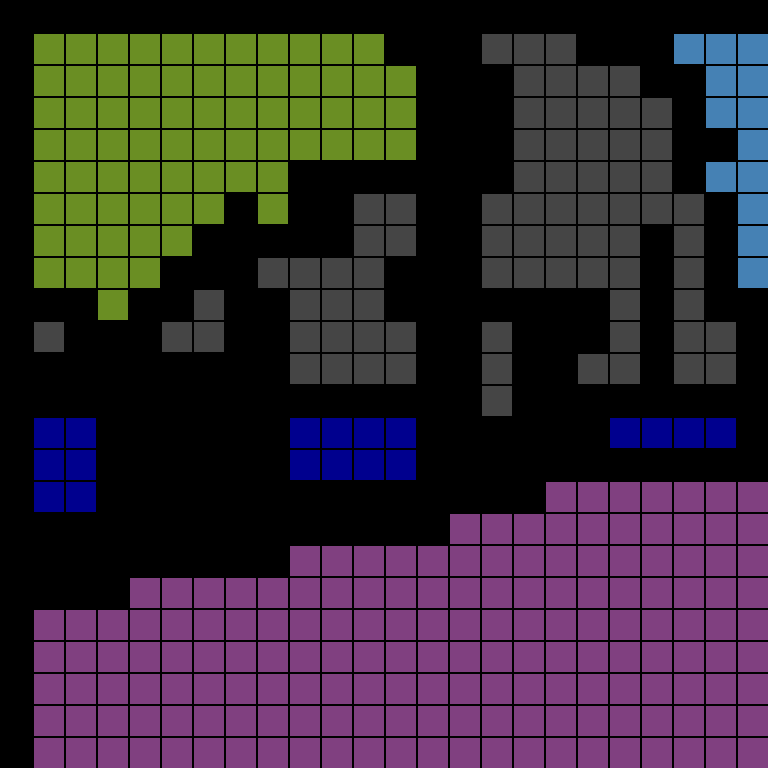}
\includegraphics[width=0.195\linewidth]{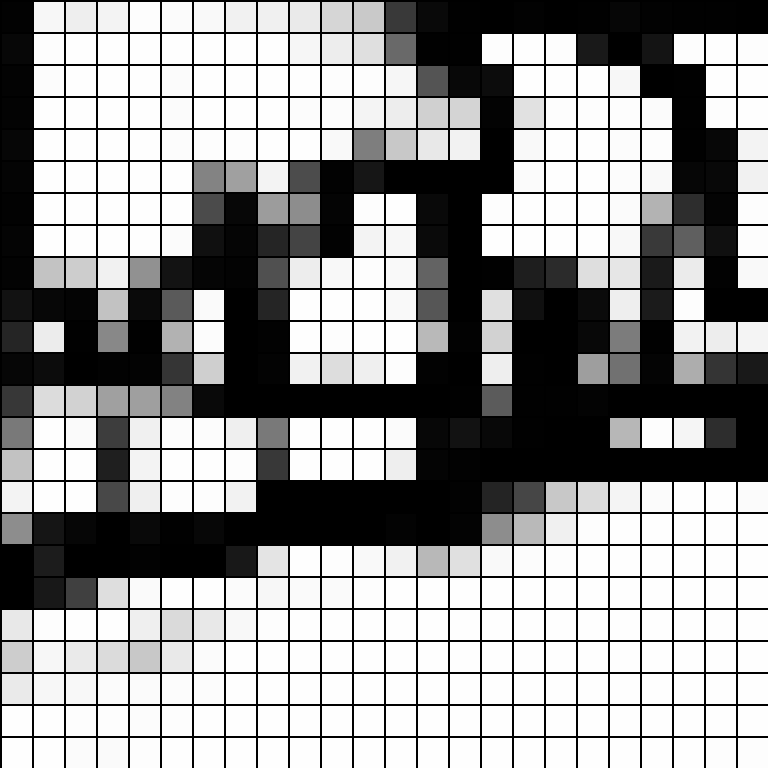}
\includegraphics[width=0.195\linewidth]{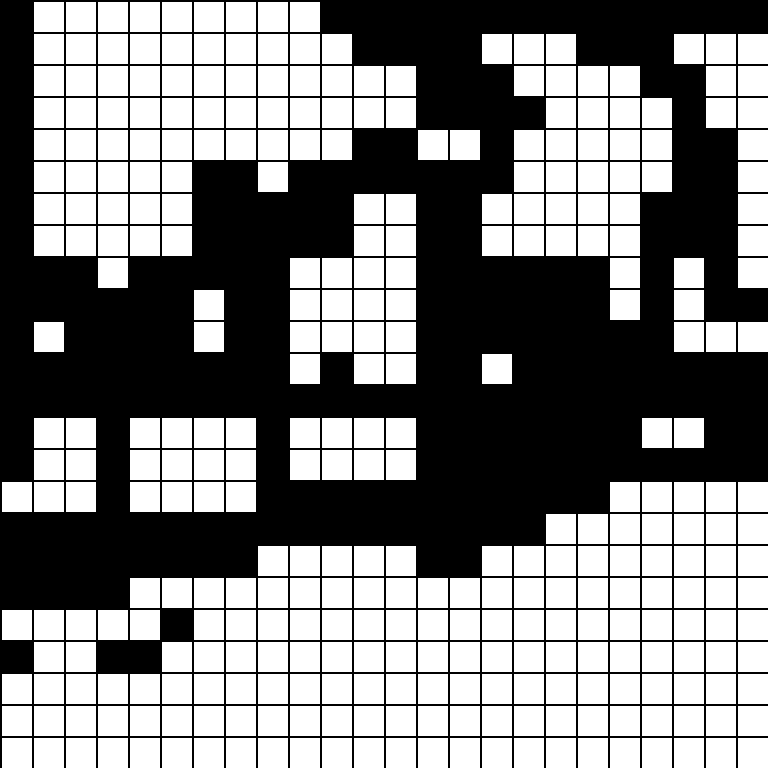} \\

\includegraphics[width=0.195\linewidth]{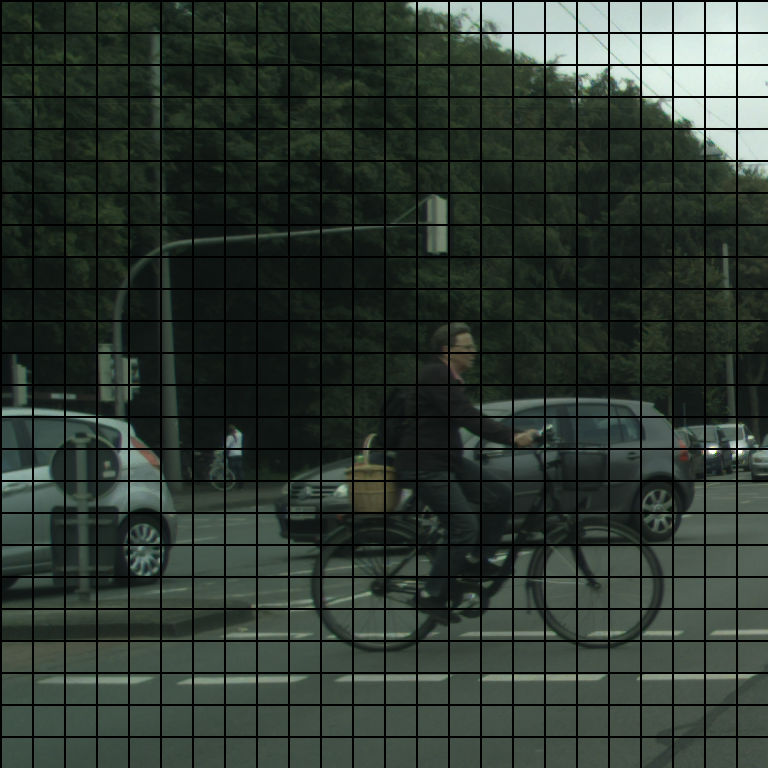}
\includegraphics[width=0.195\linewidth]{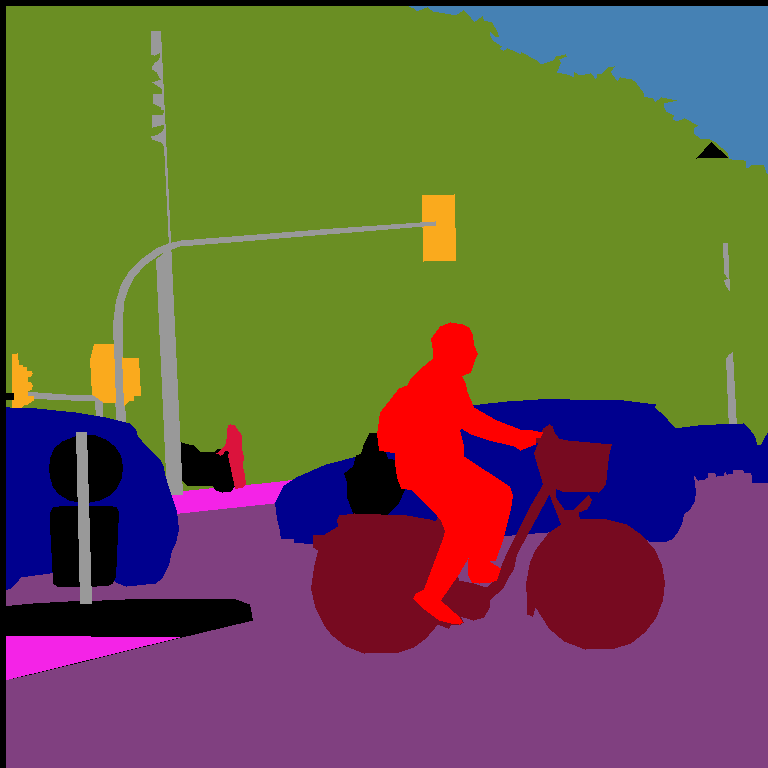}
\includegraphics[width=0.195\linewidth]{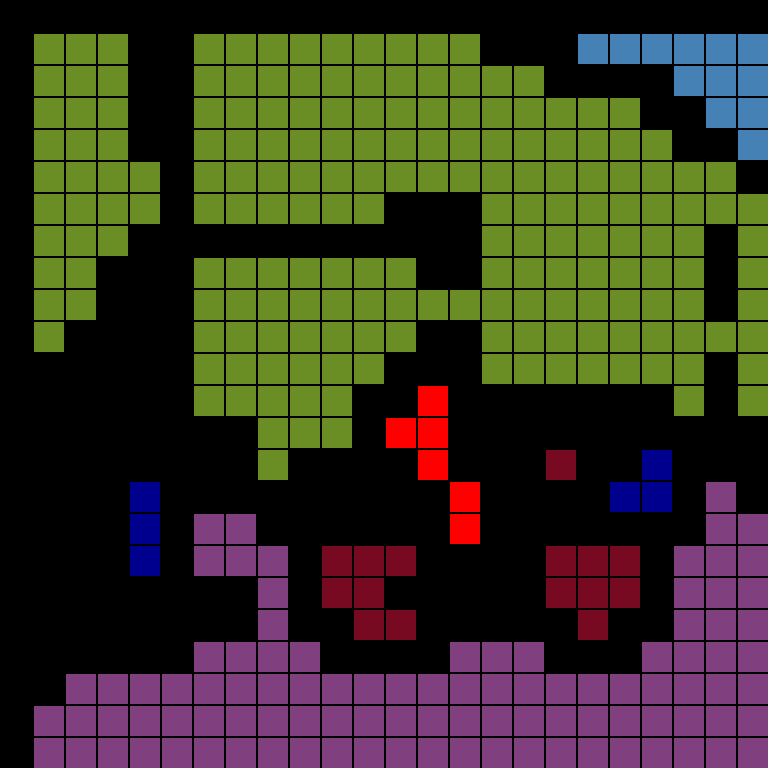}
\includegraphics[width=0.195\linewidth]{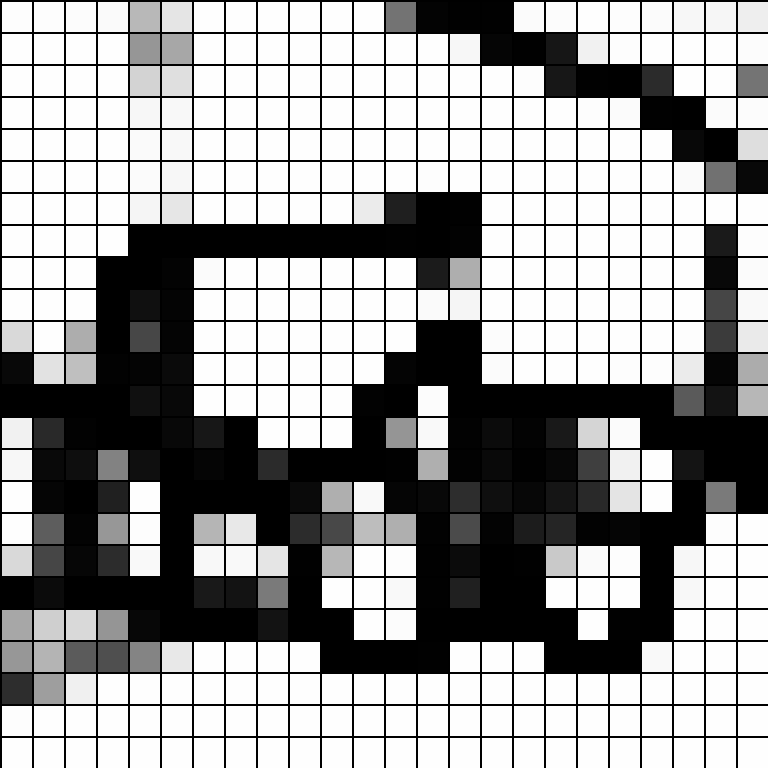}
\includegraphics[width=0.195\linewidth]{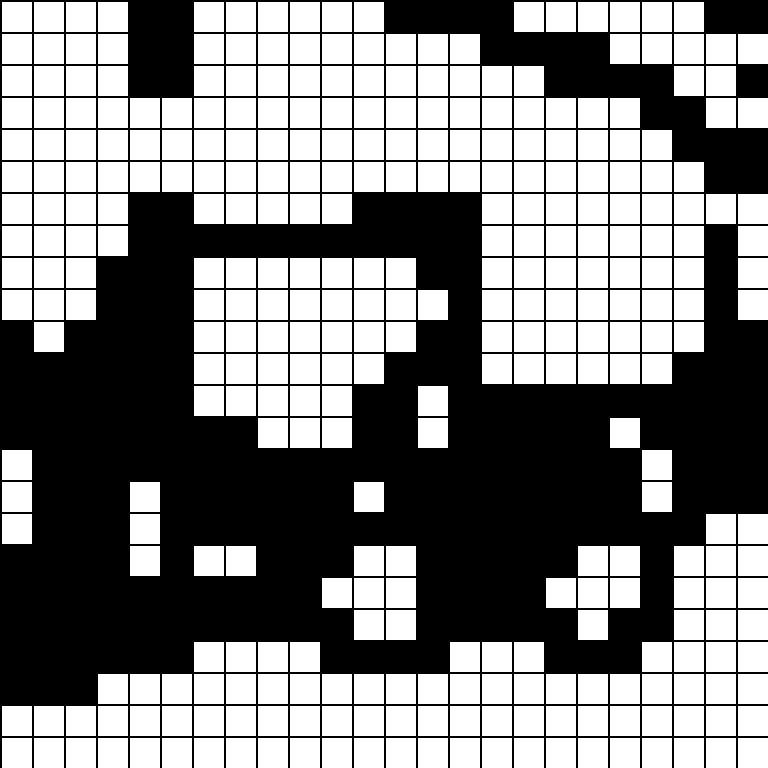} \\

\includegraphics[width=0.195\linewidth]{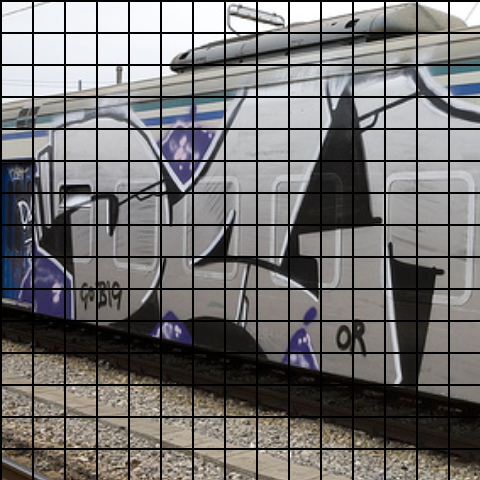}
\includegraphics[width=0.195\linewidth]{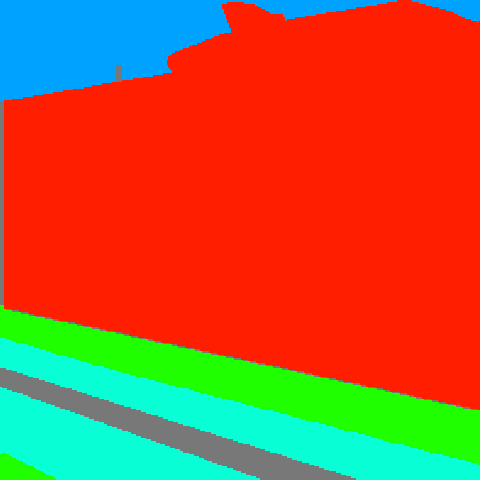}
\includegraphics[width=0.195\linewidth]{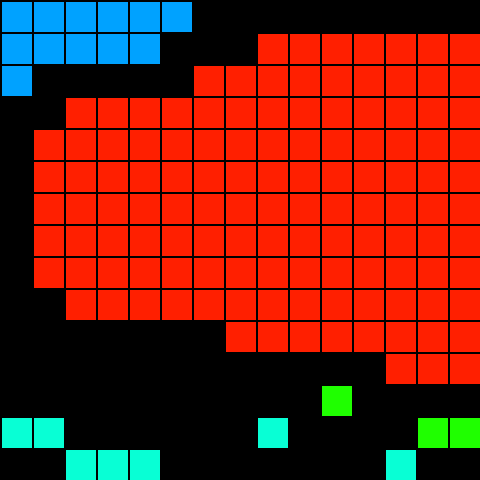}
\includegraphics[width=0.195\linewidth]{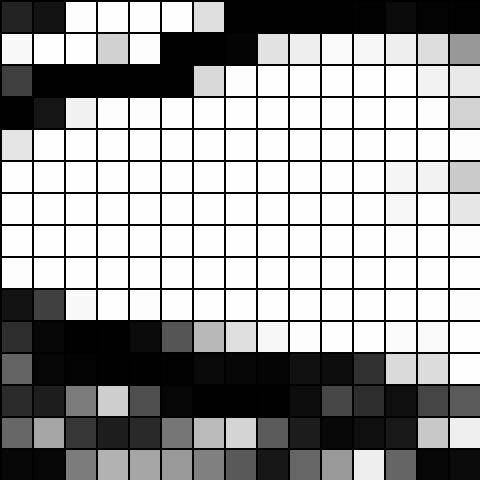}
\includegraphics[width=0.195\linewidth]{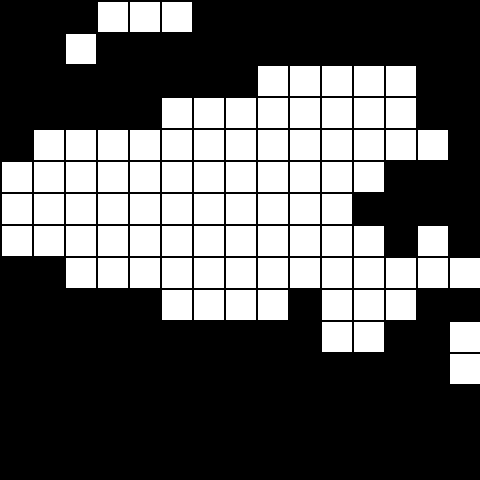}\\

\includegraphics[width=0.195\linewidth]{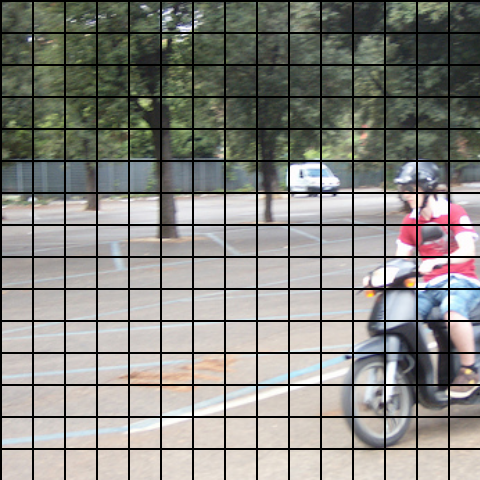}
\includegraphics[width=0.195\linewidth]{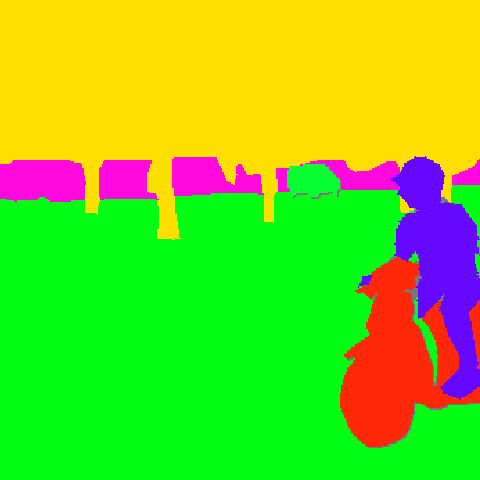}
\includegraphics[width=0.195\linewidth]{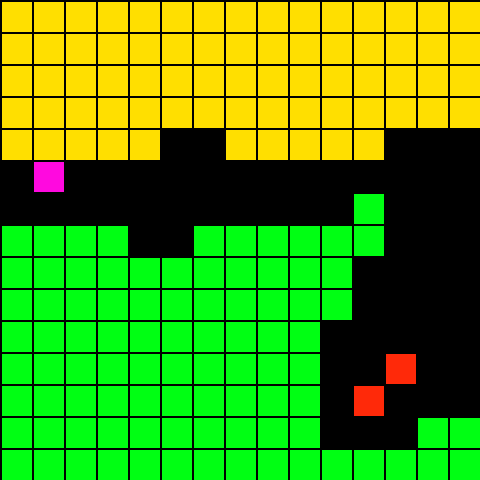}
\includegraphics[width=0.195\linewidth]{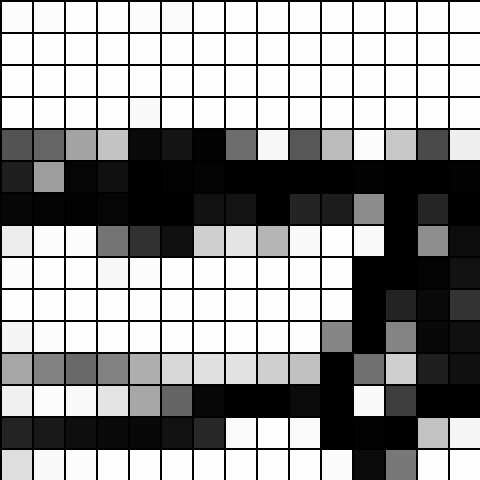}
\includegraphics[width=0.195\linewidth]{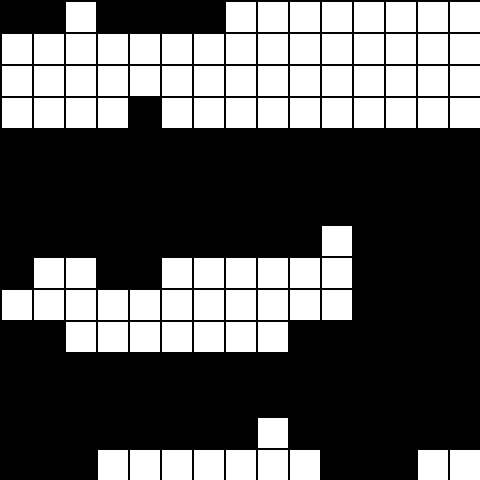}\\

\includegraphics[width=0.195\linewidth]{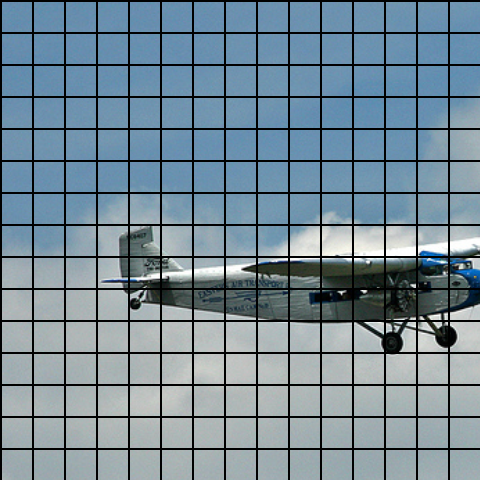}
\includegraphics[width=0.195\linewidth]{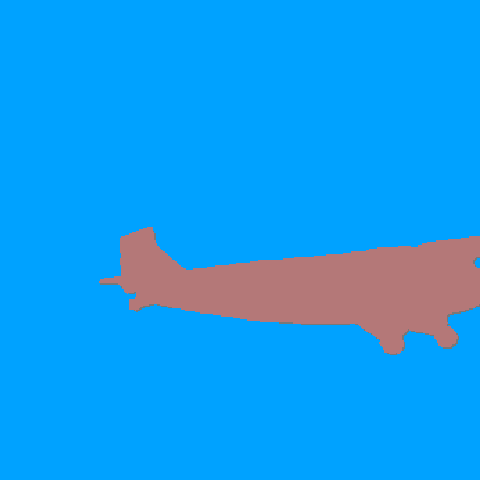}
\includegraphics[width=0.195\linewidth]{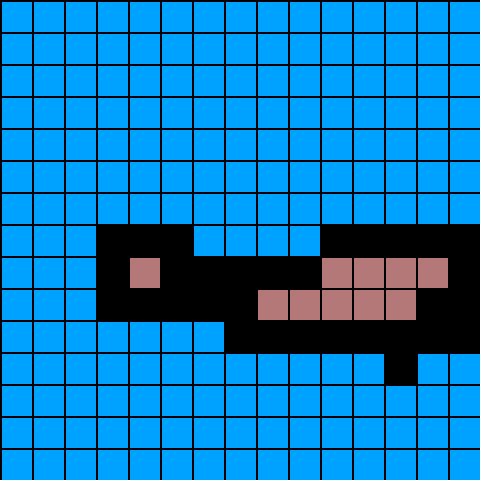}
\includegraphics[width=0.195\linewidth]{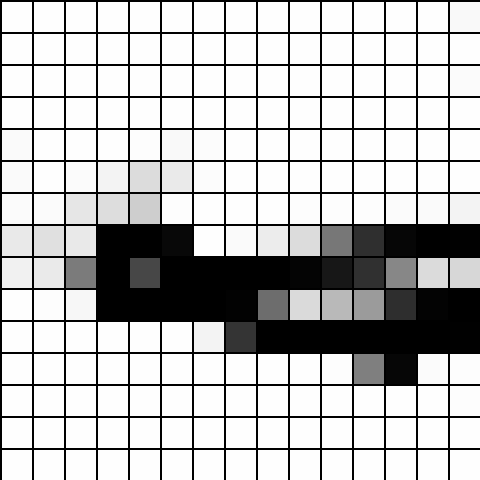}
\includegraphics[width=0.195\linewidth]{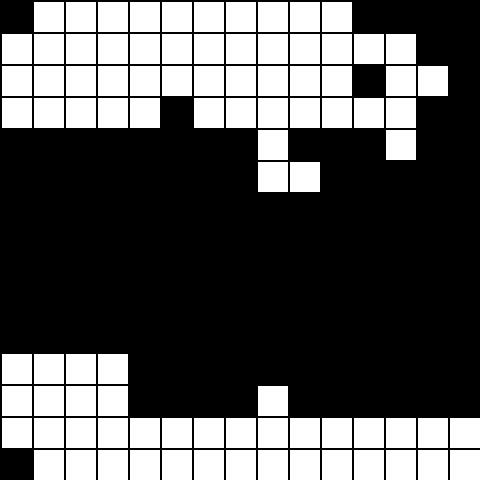}\\

 \begin{subfigure}[b]{0.195\textwidth}
     \centering
     \caption{Image with grid of superpatches}
 \end{subfigure}
 \begin{subfigure}[b]{0.195\textwidth}
     \centering
     \caption{Semantic segmentation ground truth}
 \end{subfigure}
  \begin{subfigure}[b]{0.195\textwidth}
     \centering
     \caption{Superpatches with a single class}
 \end{subfigure}
 \begin{subfigure}[b]{0.195\textwidth}
     \centering
     \caption{Policy network prediction\ \ \ \ \ \ \ \ }
 \end{subfigure}
\begin{subfigure}[b]{0.195\textwidth}
     \centering
     \caption{S highest-scoring superpatches (\ie, used policy)}
 \end{subfigure}

\caption{\textbf{Qualitative results policy network.} Predictions by our content-aware token sharing policy network, on image crops from the Cityscapes val set (top three images)~\cite{cordts_cityscapes_2016} and Pascal Context validation set (bottom three images)~\cite{mottaghi_role_2014}.} 
\label{fig:policynet_qual_results_cs_pcontext}
\end{figure*}

\paragraph{Policy network}
In \Cref{fig:policynet_qual_results_ade20k} and \Cref{fig:policynet_qual_results_cs_pcontext}, we show qualitative examples of predictions by our CTS policy network. In the figures we can see that, for all three datasets, the policy network learns to predict what superpatches contain a single semantic class quite accurately. As a result, the S highest-scoring superpatches, \ie, the superpatches that share a token, mostly consist of a single semantic class, as is the intended behavior.

\begin{figure*}[t]
\centering
\includegraphics[width=0.24\linewidth]{appendix/images/ade20k/0035_img_grid_32.png}
\includegraphics[width=0.24\linewidth]{appendix/images/ade20k/0035_policy_pred_chosen_grid.png}
\includegraphics[width=0.24\linewidth]{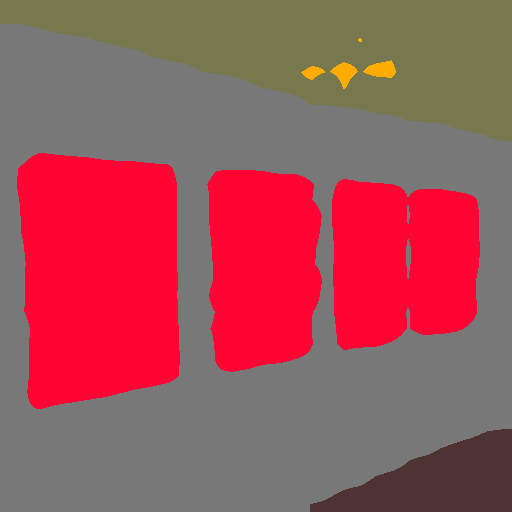}
\includegraphics[width=0.24\linewidth]{appendix/images/ade20k/0035_gt.png} \\

\includegraphics[width=0.24\linewidth]{appendix/images/ade20k/0037_img_grid_32.png}
\includegraphics[width=0.24\linewidth]{appendix/images/ade20k/0037_policy_pred_chosen_grid.png}
\includegraphics[width=0.24\linewidth]{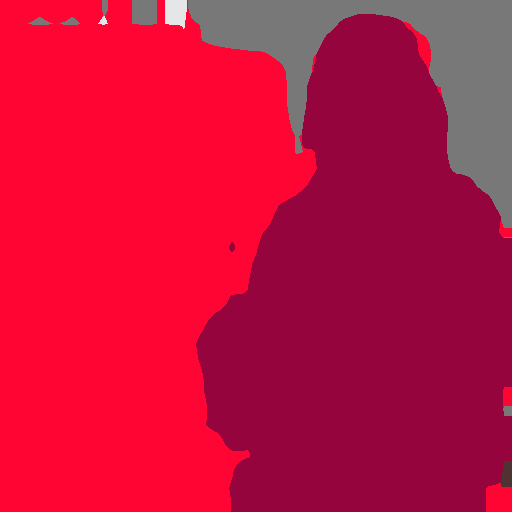}
\includegraphics[width=0.24\linewidth]{appendix/images/ade20k/0037_gt.png} \\

\includegraphics[width=0.24\linewidth]{appendix/images/ade20k/0043_img_grid_32.png}
\includegraphics[width=0.24\linewidth]{appendix/images/ade20k/0043_policy_pred_chosen_grid.png}
\includegraphics[width=0.24\linewidth]{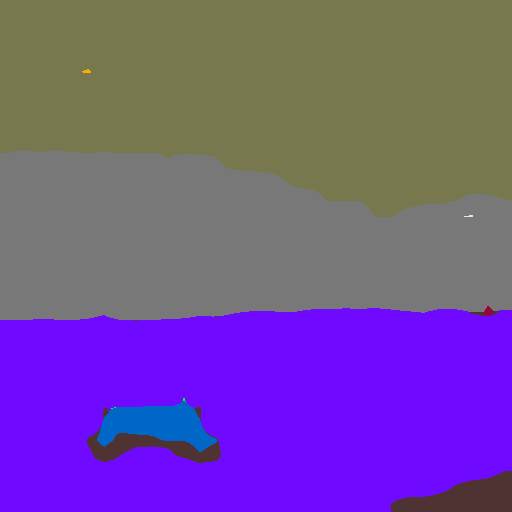}
\includegraphics[width=0.24\linewidth]{appendix/images/ade20k/0043_gt.png} \\

\includegraphics[width=0.24\linewidth]{appendix/images/ade20k/0045_img_grid_32.png}
\includegraphics[width=0.24\linewidth]{appendix/images/ade20k/0045_policy_pred_chosen_grid.png}
\includegraphics[width=0.24\linewidth]{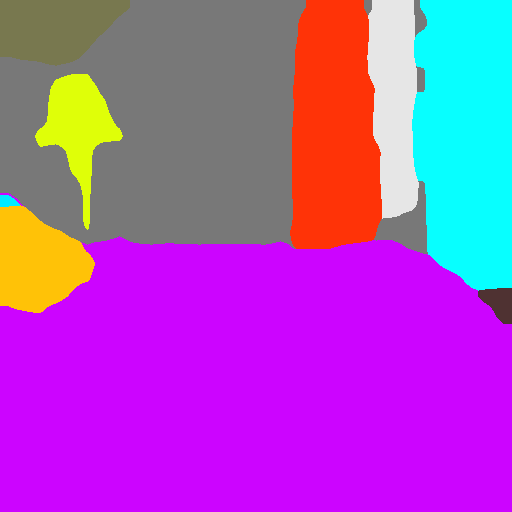}
\includegraphics[width=0.24\linewidth]{appendix/images/ade20k/0045_gt.png} \\

\includegraphics[width=0.24\linewidth]{appendix/images/ade20k/0074_img_grid_32.png}
\includegraphics[width=0.24\linewidth]{appendix/images/ade20k/0074_policy_pred_chosen_grid.png}
\includegraphics[width=0.24\linewidth]{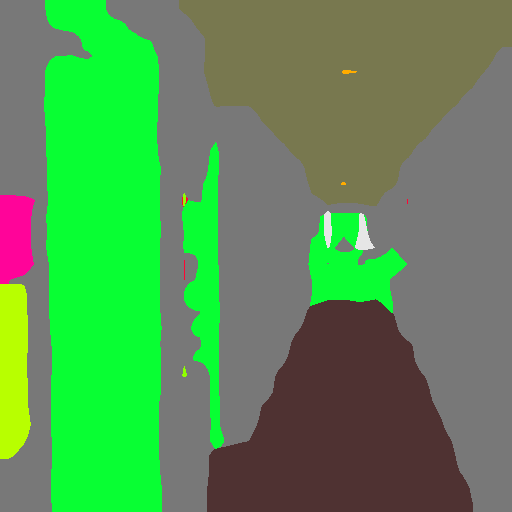}
\includegraphics[width=0.24\linewidth]{appendix/images/ade20k/0074_gt.png} \\

 \begin{subfigure}[b]{0.24\textwidth}
     \centering
     \caption{Input image}
 \end{subfigure}
 \begin{subfigure}[b]{0.24\textwidth}
     \centering
     \caption{Used token sharing policy}
 \end{subfigure}
  \begin{subfigure}[b]{0.24\textwidth}
     \centering
     \caption{Segmentation prediction}
 \end{subfigure}
   \begin{subfigure}[b]{0.24\textwidth}
     \centering
     \caption{Segmentation ground truth}
 \end{subfigure}
\vspace{-10pt}

\caption{\textbf{Qualitative results semantic segmentation from policy.} Predictions by Segmenter with ViT-S/16 and our CTS, on image crops from the ADE20K validation set~\cite{zhou_scene_2017}.} 
\label{fig:segmentation_qual_results_policy_ade20k}
\end{figure*}

\begin{figure*}[t]
\centering
\includegraphics[width=0.24\linewidth]{appendix/images/cityscapes/ff_001016_img_grid_32.png}
\includegraphics[width=0.24\linewidth]{appendix/images/cityscapes/ff_001016_policy_pred_chosen_grid.png}
\includegraphics[width=0.24\linewidth]{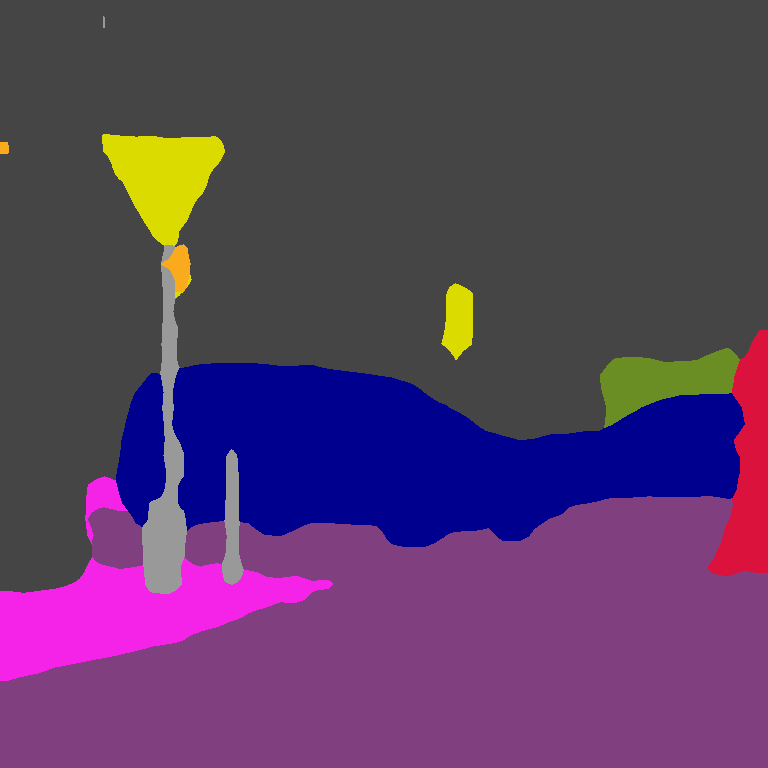}
\includegraphics[width=0.24\linewidth]{appendix/images/cityscapes/ff_001016_gt.png} \\

\includegraphics[width=0.24\linewidth]{appendix/images/cityscapes/ff_001751_img_grid_32.png}
\includegraphics[width=0.24\linewidth]{appendix/images/cityscapes/ff_001751_policy_pred_chosen_grid.png}
\includegraphics[width=0.24\linewidth]{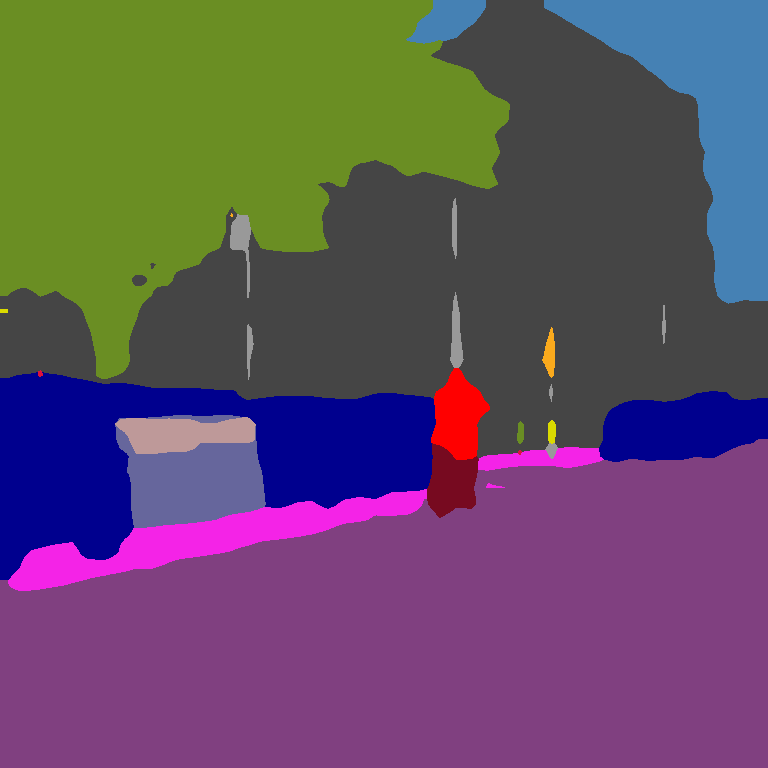}
\includegraphics[width=0.24\linewidth]{appendix/images/cityscapes/ff_001751_gt.png} \\

\includegraphics[width=0.24\linewidth]{appendix/images/cityscapes/mu_000000_img_grid_32.png}
\includegraphics[width=0.24\linewidth]{appendix/images/cityscapes/mu_000000_policy_pred_chosen_grid.png}
\includegraphics[width=0.24\linewidth]{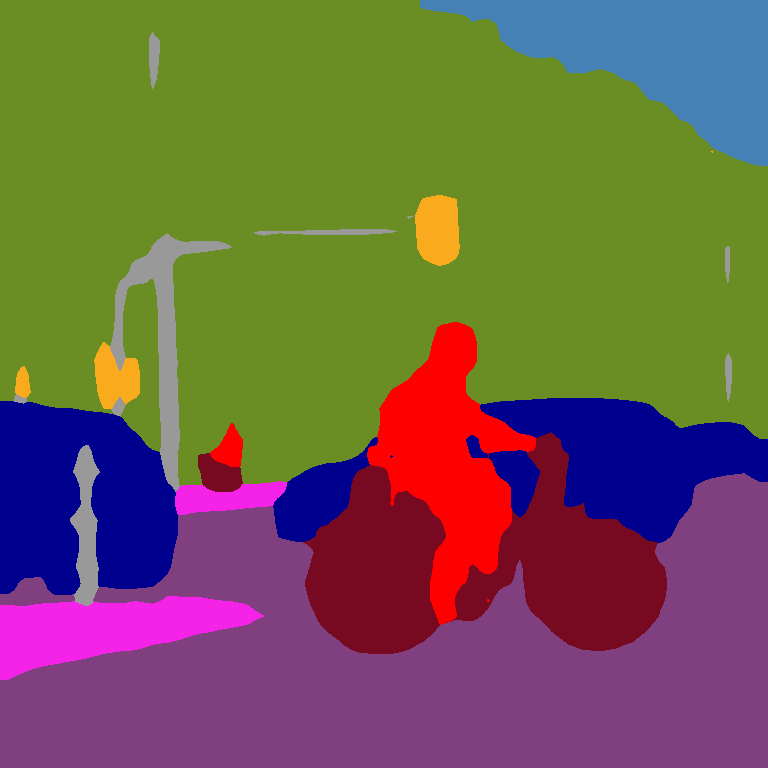}
\includegraphics[width=0.24\linewidth]{appendix/images/cityscapes/mu_000000_gt.png} \\

\includegraphics[width=0.24\linewidth]{appendix/images/pcontext/2008_000373_img_grid_32.png}
\includegraphics[width=0.24\linewidth]{appendix/images/pcontext/2008_000373_policy_pred_chosen_grid.png}
\includegraphics[width=0.24\linewidth]{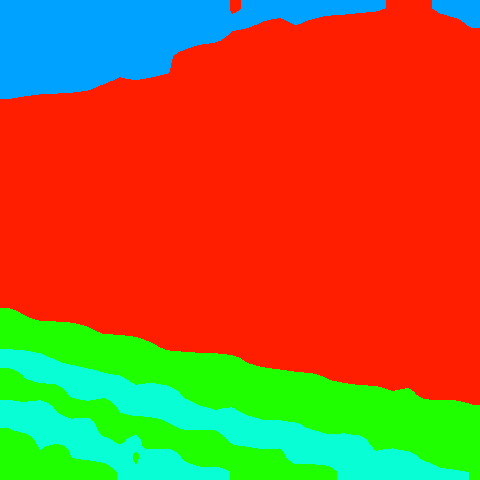}
\includegraphics[width=0.24\linewidth]{appendix/images/pcontext/2008_000373_gt.png}\\

\includegraphics[width=0.24\linewidth]{appendix/images/pcontext/2008_008127_img_grid_32.png}
\includegraphics[width=0.24\linewidth]{appendix/images/pcontext/2008_008127_policy_pred_chosen_grid.png}
\includegraphics[width=0.24\linewidth]{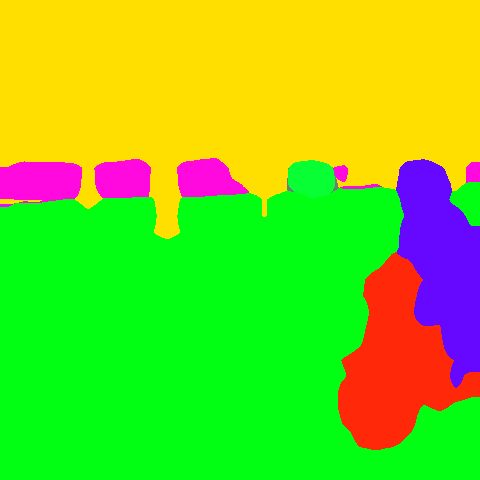}
\includegraphics[width=0.24\linewidth]{appendix/images/pcontext/2008_008127_gt.png}\\

 \begin{subfigure}[b]{0.24\textwidth}
     \centering
     \caption{Input image}
 \end{subfigure}
 \begin{subfigure}[b]{0.24\textwidth}
     \centering
     \caption{Used token sharing policy}
 \end{subfigure}
  \begin{subfigure}[b]{0.24\textwidth}
     \centering
     \caption{Segmentation prediction}
 \end{subfigure}
   \begin{subfigure}[b]{0.24\textwidth}
     \centering
     \caption{Segmentation ground truth}
 \end{subfigure}
\vspace{-10pt}

\caption{\textbf{Qualitative results semantic segmentation from policy.} Predictions by Segmenter with ViT-S/16 and our CTS, on image crops from the Cityscapes val set (top three images)~\cite{cordts_cityscapes_2016} and Pascal Context validation set (bottom two images)~\cite{mottaghi_role_2014}.} 
\label{fig:segmentation_qual_results_policy_cs_pcontext}
\end{figure*}

\paragraph{Semantic segmentation predictions}
In \Cref{fig:segmentation_qual_results_policy_ade20k} and \Cref{fig:segmentation_qual_results_policy_cs_pcontext}, we show qualitative examples of semantic segmentation predictions by Segmenter~\cite{strudel_segmenter_2021} with our CTS. These figures show that the network mostly predicts a single class for the superpatches that share a token, which is intended behavior. Additionally, we do not observe any qualitative artifacts in the predictions that result from sharing tokens.

\begin{figure*}[t]
\centering
\includegraphics[width=0.25\linewidth]{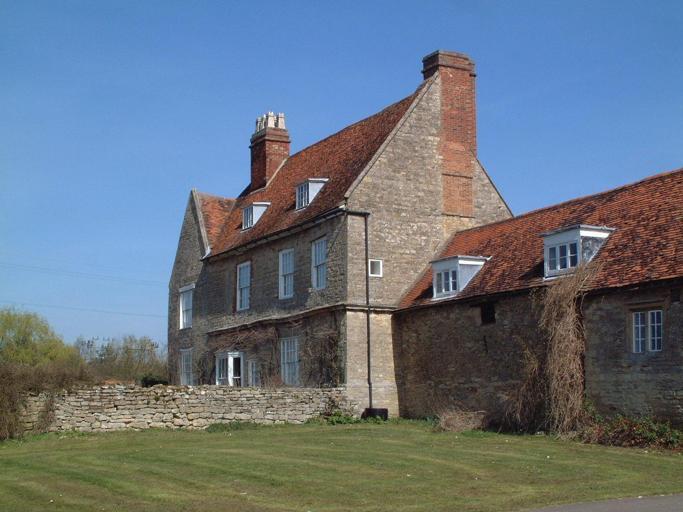}
\includegraphics[width=0.25\linewidth]{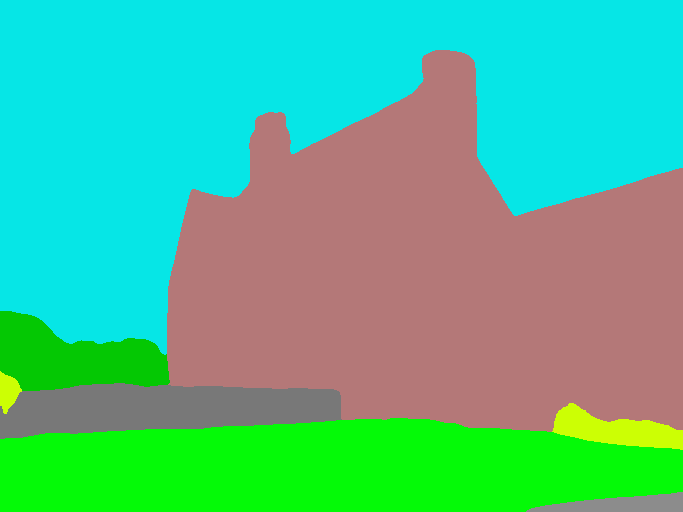}
\includegraphics[width=0.25\linewidth]{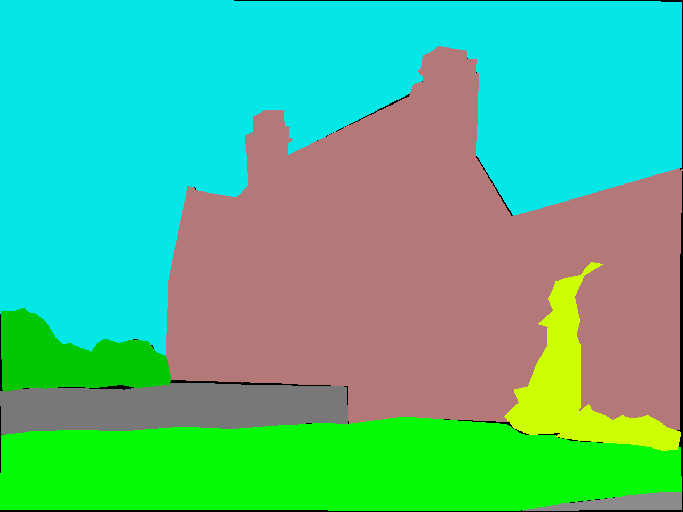} \\

\includegraphics[width=0.25\linewidth]{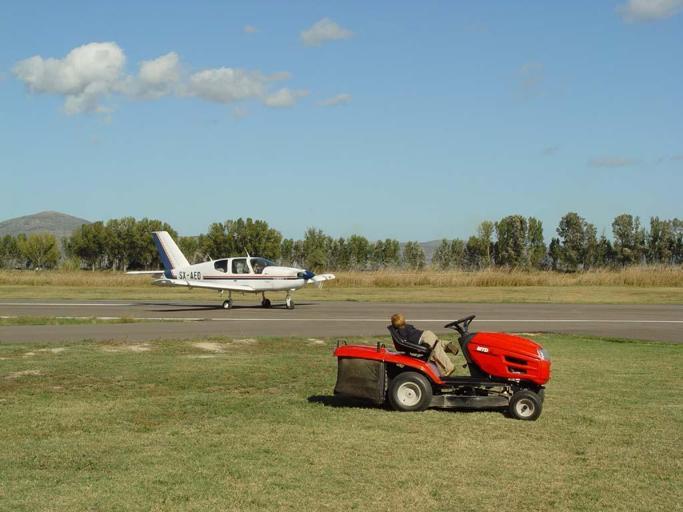}
\includegraphics[width=0.25\linewidth]{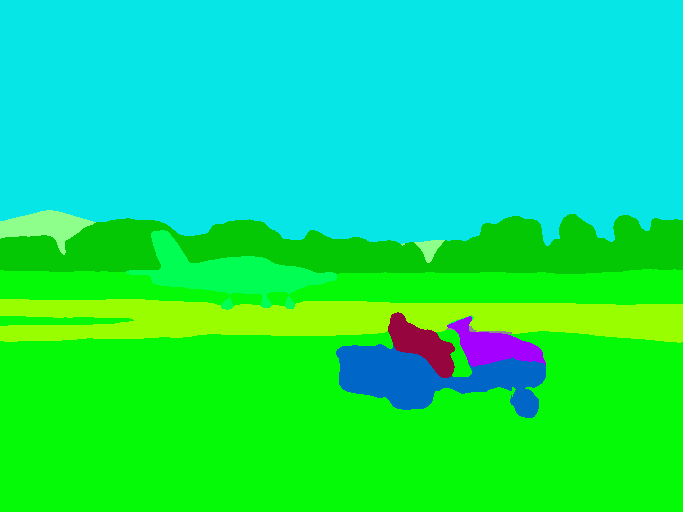}
\includegraphics[width=0.25\linewidth]{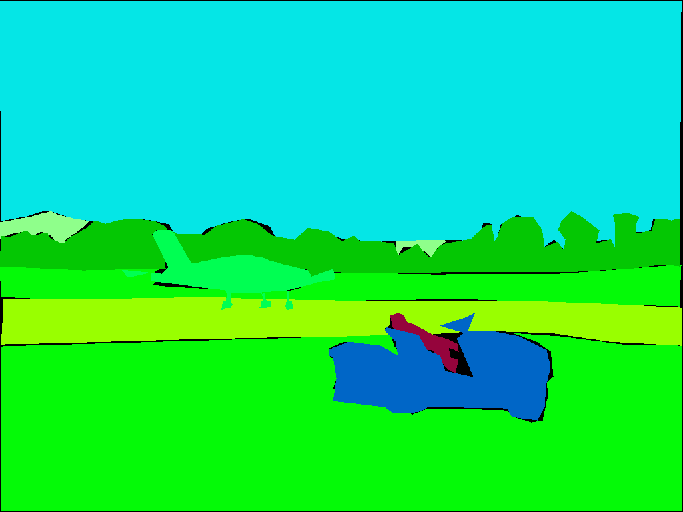} \\

\includegraphics[width=0.25\linewidth]{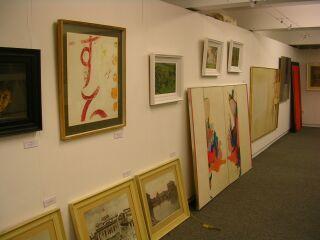}
\includegraphics[width=0.25\linewidth]{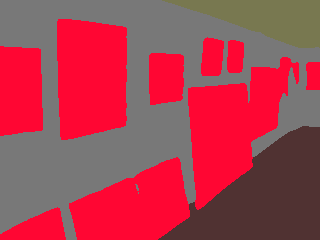}
\includegraphics[width=0.25\linewidth]{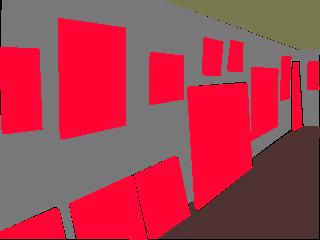} \\

\includegraphics[width=0.25\linewidth]{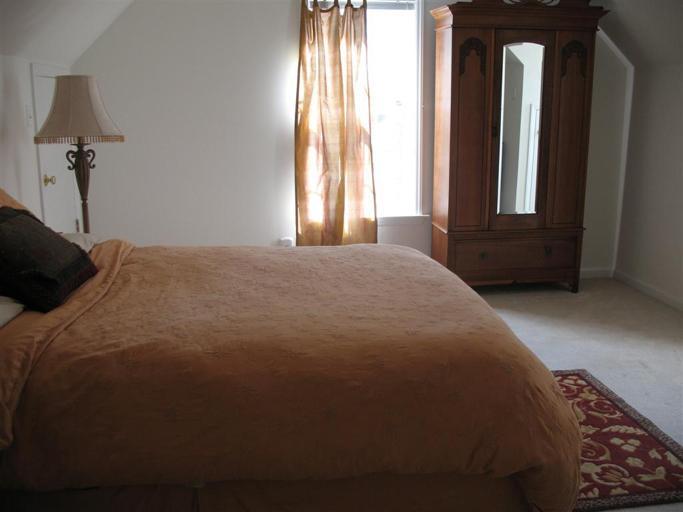}
\includegraphics[width=0.25\linewidth]{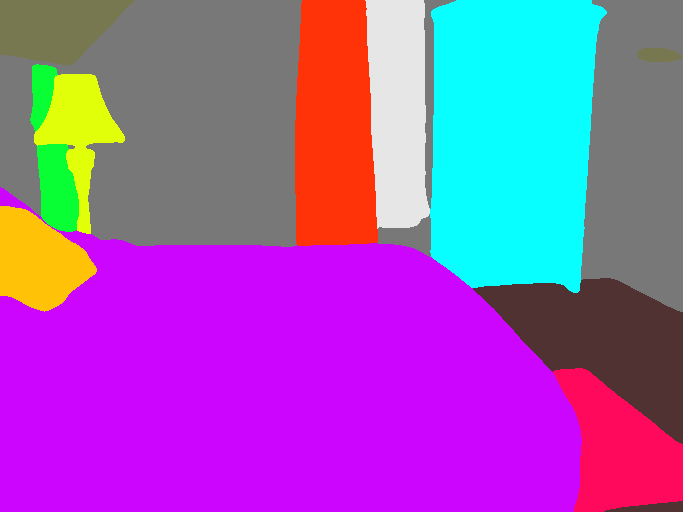}
\includegraphics[width=0.25\linewidth]{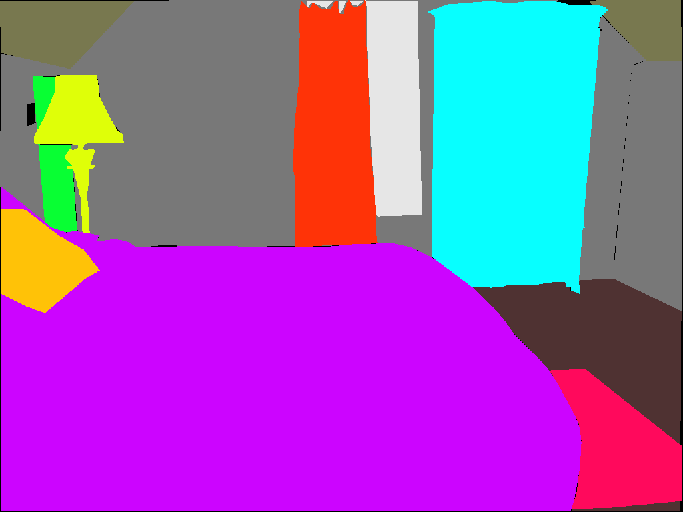} \\

\includegraphics[width=0.25\linewidth]{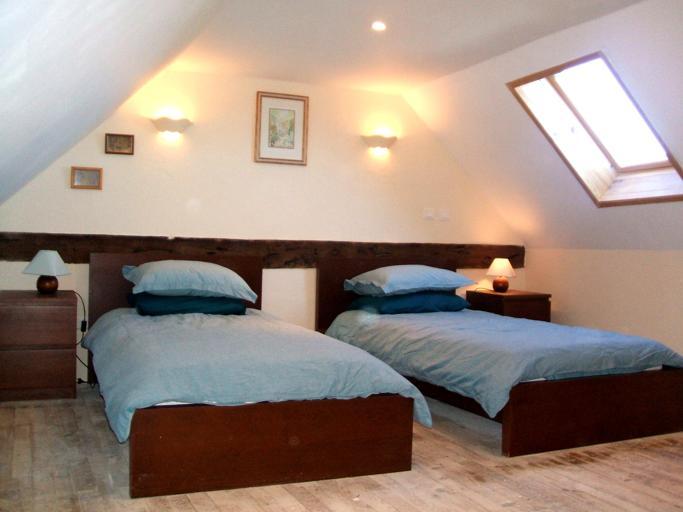}
\includegraphics[width=0.25\linewidth]{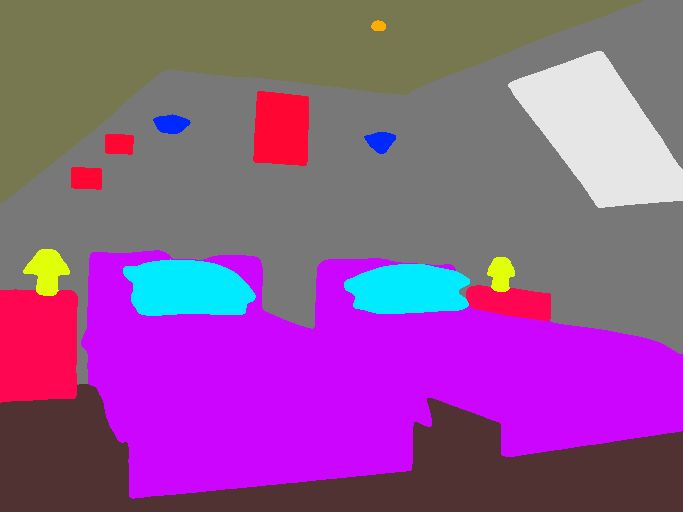}
\includegraphics[width=0.25\linewidth]{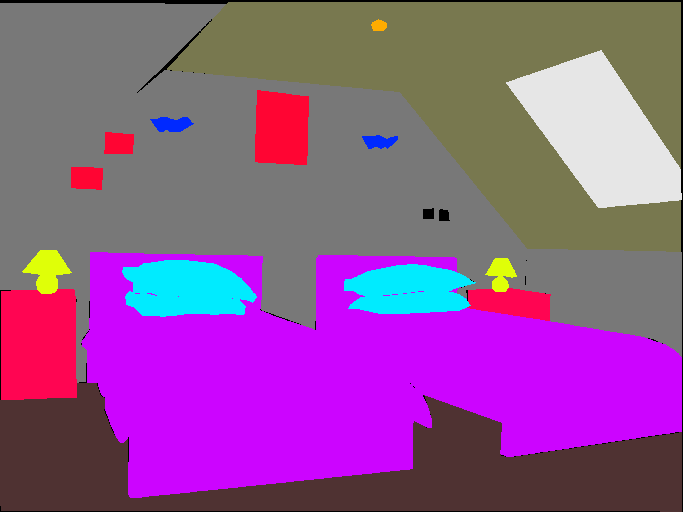} \\

\includegraphics[width=0.25\linewidth]{appendix/images/sota/ADE_val_00000086.jpg}
\includegraphics[width=0.25\linewidth]{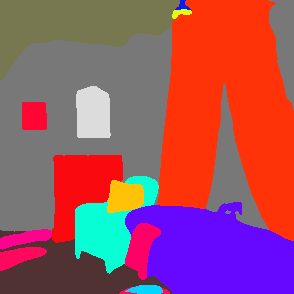}
\includegraphics[width=0.25\linewidth]{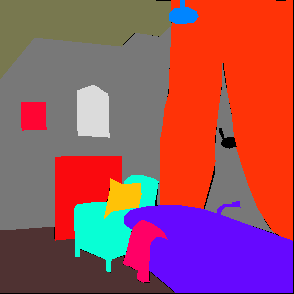} \\

 \begin{subfigure}[b]{0.25\textwidth}
     \centering
     \caption{Input image}
 \end{subfigure}
  \begin{subfigure}[b]{0.25\textwidth}
     \centering
     \caption{Segmentation prediction}
 \end{subfigure}
   \begin{subfigure}[b]{0.25\textwidth}
     \centering
     \caption{Segmentation ground truth}
 \end{subfigure}
\vspace{-10pt}

\caption{\textbf{Qualitative results semantic segmentation state-of-the-art.} Examples of results by BEiT-L + UPerNet~\cite{xiao2018unified,bao2022beit}, a state-of-the-art semantic segmentation approach, with our CTS, on the ADE20K validation set~\cite{zhou_scene_2017}. With our CTS, we are able to increase the throughput of this network by more than 64\%, without decreasing the segmentation quality. } 
\label{fig:segmentation_qual_results_sota}
\end{figure*}

Finally, we show the results of our CTS with state-of-the-art semantic segmentation network BEiT-L + UPerNet~\cite{bao2022beit,xiao2018unified} in \Cref{fig:segmentation_qual_results_sota}. These results show that it is possible to achieve state-of-the-art segmentation quality with our CTS, while also achieving a throughput increase of over 60\%.

\end{document}